\documentclass{article}
\usepackage{log_2023}						

\usepackage{booktabs}						
\usepackage{multirow}						
\usepackage{amsfonts}						
\usepackage{graphicx}						
\usepackage{duckuments}						

\usepackage[numbers,compress,sort]{natbib}	

\title[ Generative Modeling of Labeled Graphs under Data Scarcity]{Generative Modeling of Labeled Graphs under Data Scarcity}

\author{
  Sahil Manchanda\thanks{Equal contribution.} \And
  Shubham Gupta\footnotemark[1] \And
  Sayan Ranu\\
\\\And
  Srikanta Bedathur \AND
Department of Computer Science and Engineering \\
Indian Institute of Technology Delhi\\
  \email{\{sahil.manchanda,shubham.gupta,sayanranu, srikanta\}@cse.iitd.ac.in} \\
}

\usepackage{soul}
 \usepackage{epstopdf}
\usepackage[utf8]{inputenc}
\usepackage{url,textcomp}  
\usepackage{tablefootnote}
\usepackage{mdwlist}
\usepackage{tabularx}
\usepackage{amsmath,amsfonts,amsthm,bm}
\usepackage{graphicx}
\usepackage{enumitem}
\usepackage{array}
\usepackage{arydshln}
\usepackage{mathtools}
\usepackage{flushend}  

\usepackage{xspace}
\usepackage{xcolor}
\usepackage{graphicx}
\usepackage{amsfonts}
\usepackage{booktabs}

\usepackage{stmaryrd}

\def\GraphGen{\textsc{GraphGen}\xspace}

\usepackage{algorithmic}
\usepackage[ ruled,vlined, norelsize]{algorithm2e}

\usepackage{todonotes}
\usepackage[toc,page]{appendix}
\usepackage{diagbox}
\usepackage{multirow}
\usepackage{array}
\usepackage{makecell}
\usepackage{fmtcount} 
\usepackage{array,multirow}
\usepackage{tablefootnote}
\usepackage{makecell}

  \usepackage{graphicx}

\usepackage{subfig}
\usepackage{caption}

 \setlist{nolistsep,leftmargin=*}

\newcommand{\ch}{\mathbf{h}\xspace}

\usepackage{amsfonts}

 \newtheorem{defn}{\textbf{Definition}}
  \newtheorem{prob}{\textbf{Problem}}

\def\namemodel{\textsc{GShot}\xspace}
\def\graphRNN{\textsc{GraphRNN}\xspace}
\def\graphGen{\textsc{GraphGen}\xspace}

\newcommand{\rev}[1]{\textcolor{black}{#1}}

\begin{document}
\maketitle

\begin{abstract}
 \vspace{-0.05in}  Deep graph generative modeling has gained enormous attraction in recent years due to its impressive  ability to directly learn the underlying hidden graph distribution. 
Despite their initial success, these techniques, like \rev{many} of the existing deep generative methods, require a large number of training samples to learn a good model. Unfortunately, \rev{a}  large number of training samples may not always be available in scenarios such as drug discovery for rare diseases. At the same time, recent advances in few-shot learning have opened door to applications where available training data is limited. In this work, we introduce the hitherto unexplored paradigm of \textit{\rev{labeled} graph generative modeling under data scarcity.} Towards this, we develop \textsc{GShot}, a meta-learning based framework for \textit{labeled} graph generative modeling under data scarcity. \textsc{GShot} learns to  transfer meta-knowledge from similar auxiliary graph datasets. Utilizing these prior experiences, \textsc{GShot} quickly adapts to an unseen graph dataset through \textit{self-paced fine-tuning}. Through extensive experiments on datasets from diverse domains having limited training samples, we establish that \textsc{GShot} generates graphs of superior fidelity compared to existing baselines.

\end{abstract}

\vspace{-0.20in}
\section{Introduction and Related Work}
\label{sec:intro}
\vspace{-0.10in}


With recent advances in deep learning, there has been a surge in developing deep graph generative methods that directly learn the underlying hidden distribution of graphs from the data itself \cite{you2018graph,goyal2020graphgen,molgan,netgan,albert2002statistical,gran,graphvae,gupta2022tigger,gupta2022survey,neuromlr}. These techniques have shown significant improvement over the traditional methods for the graph generation task. 
Since many real-world graphs such as protein interaction networks \cite{bogwartenzymes} and drug molecules \cite{cancer} are labeled and originate from diverse domains, our focus is on learning \textit{domain-agnostic}, \textit{labeled} graph generative~\cite{goyal2020graphgen,graphrnn} model which jointly models the relationships between a graph structure and its node/edge labels. 

\begin{figure}[t]
\vspace{-0.35in}
\centering
\hspace{-0.15in}
\subfloat[\textbf{Performance vs Number of training samples \label{fig:lung-perf-det}}]{
\hspace{-0.15in}\includegraphics[scale=0.16]{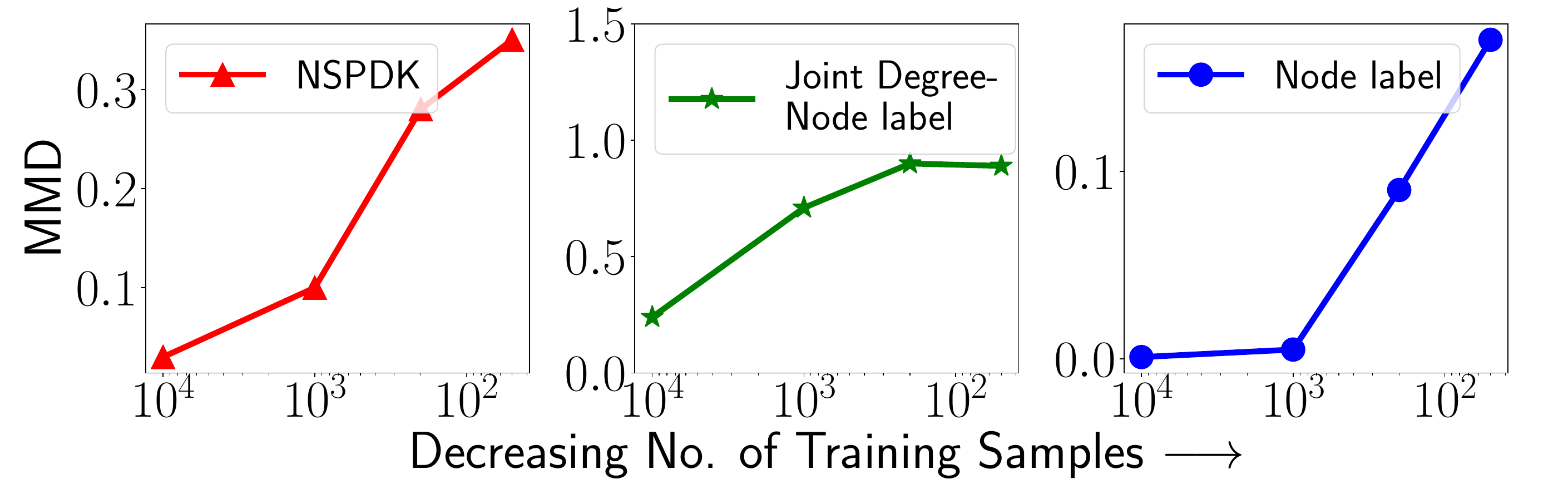}} \hspace{0.05in} \subfloat[ The pipeline of steps in  \namemodel.\label{fig:framework}]{
\includegraphics[scale=0.145]{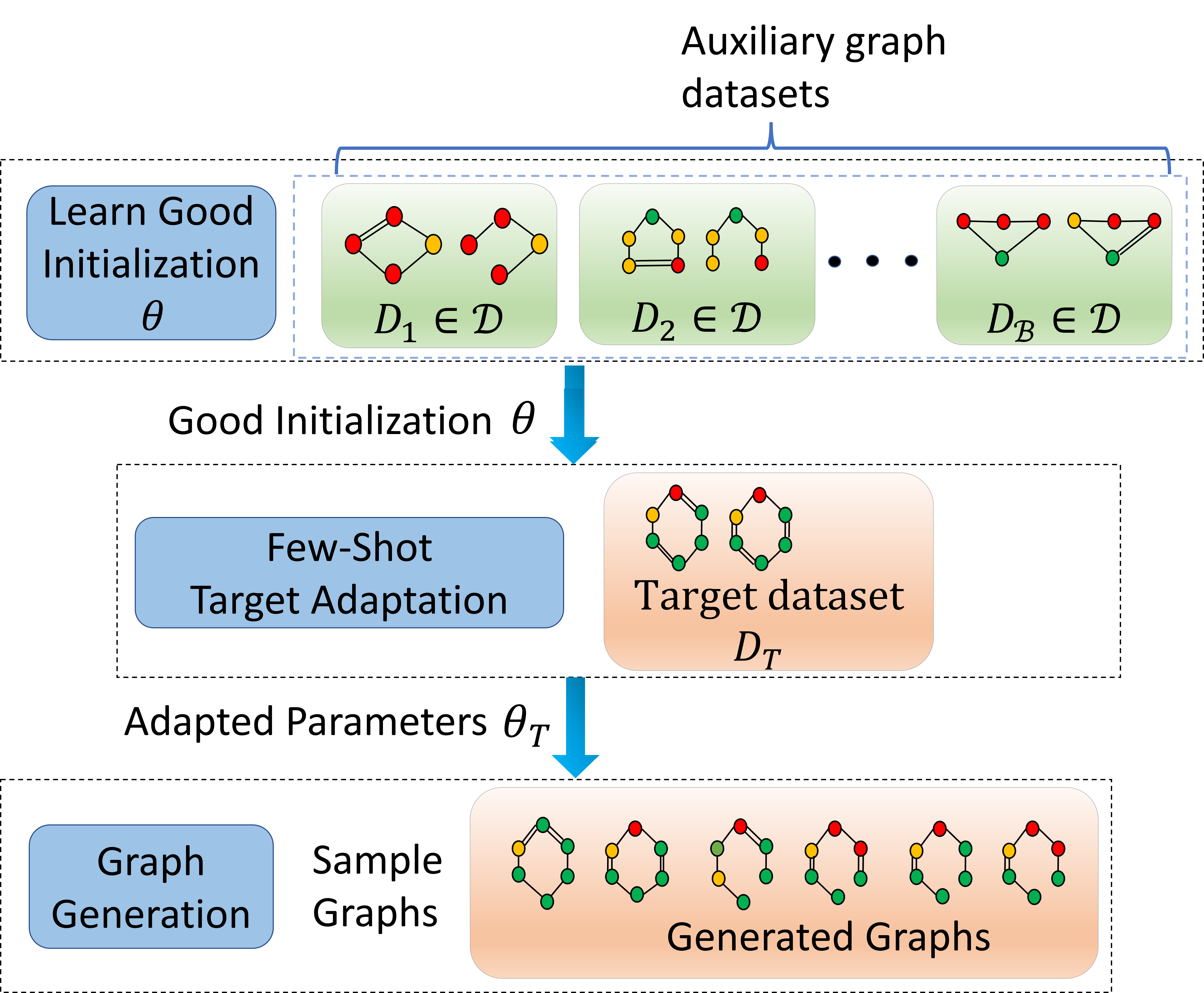}
  }
 \caption{\textbf{(a)} Increase in Maximum Mean Discrepancy (MMD) scores for different graph metrics when the number of training samples (log scale) are decreased in a chemical compound dataset (Dataset \#2 in Table~\ref{tab:datasets}) for \graphGen \cite{goyal2020graphgen}. A higher MMD corresponds to poor fidelity. \textbf{(b)} Our proposed architecture.}
\label{fig:}
\vspace{-0.25in}
\end{figure}

A well-known fact about deep generative models is that they are not well suited for applications where training data is scarce~\cite{bartunov2018few}. In our study, we observe similar trends for graph deep generative modeling. In Fig. ~\ref{fig:lung-perf-det} we study the impact of limiting the number of training samples available to \GraphGen~\cite{goyal2020graphgen}, which is the  state-of-the-art method for domain agnostic labeled graph generation. We observe that \graphGen's performance deteriorates significantly\footnote{See Sec.~\ref{sec:metrics} for detailed understanding of the metrics.} when the size of the training dataset is reduced.

The lack of training graphs is often severe in many important settings such as effective drug discovery for rare diseases~\cite{swinney2014discovery} or speedy drug discovery during pandemics such as \textsc{Covid}-19~\cite{10.3389/fmolb.2020.616341}.  Similar issue appears in physics while developing generative models for computationally expensive \textit{N-body} simulations~\cite{kipf2018neural, perraudin2019cosmological, du2021graphgt}.

In this context, we observe that although the availability of graphs exhibiting a specific desired property may be limited, it may be possible to identify graph repositories exhibiting similar properties. To elaborate, we may not have access to a large set of molecules exhibiting activity against \textsc{Covid}-19. However, million-scale repositories of chemical compounds are widely available~\cite{irwin2005zinc}, from which the broad characteristics of chemical compounds such as valency rules, correlated functional groups, etc. may be learned. Hence, potentially, the learning task from the smaller \textsc{Covid}-19 repository could be focused only on features that are unique to this set. 
We exploit this intuition and make the \rev{following novel contributions}:

\rev{
\begin{itemize}
\item \textbf{Problem Formulation:} We formulate the problem of  \textit{domain-agnostic}, \textit{labeled} graph generative modeling  \textit{under data scarcity}. To the best of our knowledge, we are the first to investigate this problem. 
\item \textbf{Algorithm:} We propose \namemodel, that effectively integrates  meta-learning with auto-regressive graph generative model to facilitate transfer of knowledge from a set of auxiliary graph datasets to target graph dataset(s). Meta-learning on these auxiliary datasets fosters a high quality parameter initialization. Subsequently, using a \textit{self-paced} fine-tuning approach, \namemodel adapts to unseen target graph dataset using a small number of training samples.  
\item \textbf{Empirical Evaluation:} We perform extensive  experiments across multiple real labeled graph datasets spanning a variety of domains such as chemical compounds,  proteins, and physical interaction systems. We establish that \namemodel is effective in learning graph distributions with high fidelity even on datasets containing as few as $50$ training samples, and significantly improves over baselines that learn from scratch. 
\end{itemize}
}

\vspace{-0.15in}
\section{Problem Formulation \protect\footnote{All notations used in our work are summarized in Table \ref{tab:notation} in the appendix.}}
\label{sec:formulation}
\vspace{-0.05in}

\noindent
\begin{defn}[Graph]
\label{def:graph}
A graph is represented as $G = (V, E)$, where  $V = \{v_{1},\cdots, v_{n}\}$ is a set of $n$ nodes and $E = \{(v_{i}, v_{j}) \mid v_{i}, v_{j} \in V \}$ is a set of edges.
 Let $\mathbb{L}_{node}:V \to \mathbb{V}$  and $\mathbb{L}_{edge}: E \to \mathbb{E} $ be the node and edge label mappings respectively where $\mathbb{V}$ and $\mathbb{E}$ are the set of all node and edge labels respectively. We assume that the graph is connected and there are no self-loops.
\end{defn}

A \textit{graph dataset} $D{=}\{G_1,\cdots, G_N\}$ is a collection of $N$ graphs. Graph dataset $D_1$ is considered to be an \textit{auxiliary} dataset of graph dataset $D_2$ if $D_1$ is \textit{similar} to $D_2$. As discussed in Sec.~\ref{sec:intro}, a generic set of chemical compounds may be considered as an auxiliary dataset to a specific subgroup of compounds that display a desired activity against a virus. \rev{In the current context of \textit{labeled domain-agnostic} graph generative modeling, we below define the concept of \textit{graph dataset distance metric} and subsequently formally define \textit{auxiliary dataset}.}
\rev{ 
\begin{defn}[Graph Dataset Distance Metric] A graph dataset distance metric is a function $\Lambda_m(D_i, D_j)\mapsto \mathbb{R}$ that takes two graph datasets $D_i$ and $D_j$ as input and computes distance between the two datasets as per metric $\Lambda_m$. \label{def:graph_dist_metric}
\end{defn} Several metrics have been proposed in the literature for comparing graph datasets~\cite{goyal2020graphgen, graphrnn}. In Sec.~\ref{sec:eval_metrics} we describe various distance metrics.}
\rev{
\begin{defn}[Auxiliary Dataset] A graph dataset is considered to be an auxiliary dataset to a target dataset if it satisfies the following conditions:
\begin{enumerate}
    \item The node/edge label space of the target and auxiliary datasets should be similar.
    \item The distances between auxiliary and target graph datasets should be low as per Defn.~\ref{def:graph_dist_metric}.\footnote{\rev{In App Sec.~\ref{app:sim_tar_aux} we present similarity results between datasets on several metrics.} }
\end{enumerate}
\label{def:aux_dataset}
\end{defn}
}

\begin{prob}[Graph Generative modelling]
The goal of labeled graph\footnote{In our paper we use the keyword \textit{graph} and \textit{labeled graph} interchangeably} generative modeling of a dataset $D$ of graphs is to learn a model, ${p}_\theta (D)$, parameterized by $\theta$, that approximates the true latent distribution $p(D)$ of graphs in $D$. The learned generative model is effective if it is capable of generating graphs similar to those in $D$.
\end{prob}

The goal is to learn a generative model over a target graph dataset $D_T$, where $|D_T|$ is small. Since $|D_T|$ is small, accurate modeling is hard (Recall Fig.~\ref{fig:lung-perf-det}). However, if $D_T$ is accompanied with a collection of auxiliary datasets, the generative model should be able to use this knowledge and augment its learning. Formally, it is defined as follows.
\noindent
\begin{prob}[Labeled Graph Generative Modelling under Data Scarcity]\hfill\newline
\noindent
\textbf{Input:} 
A collection of auxiliary graph datasets $\mathcal{D}{=}\{D_1,\cdots, D_{\mathcal{B}} \}$ and a target dataset $D_T$.\hfill\newline
\noindent
\textbf{Goal:} To learn a graph generative model $p_{\theta}(D)$ that is capable of leveraging the knowledge from $\mathcal{D}$ and effectively \rev{adapting} to the unseen target dataset $D_T$.
\end{prob}

\begin{figure*}[t!]
\vspace{-0.20in}
\centering
\includegraphics[width=0.9\textwidth]{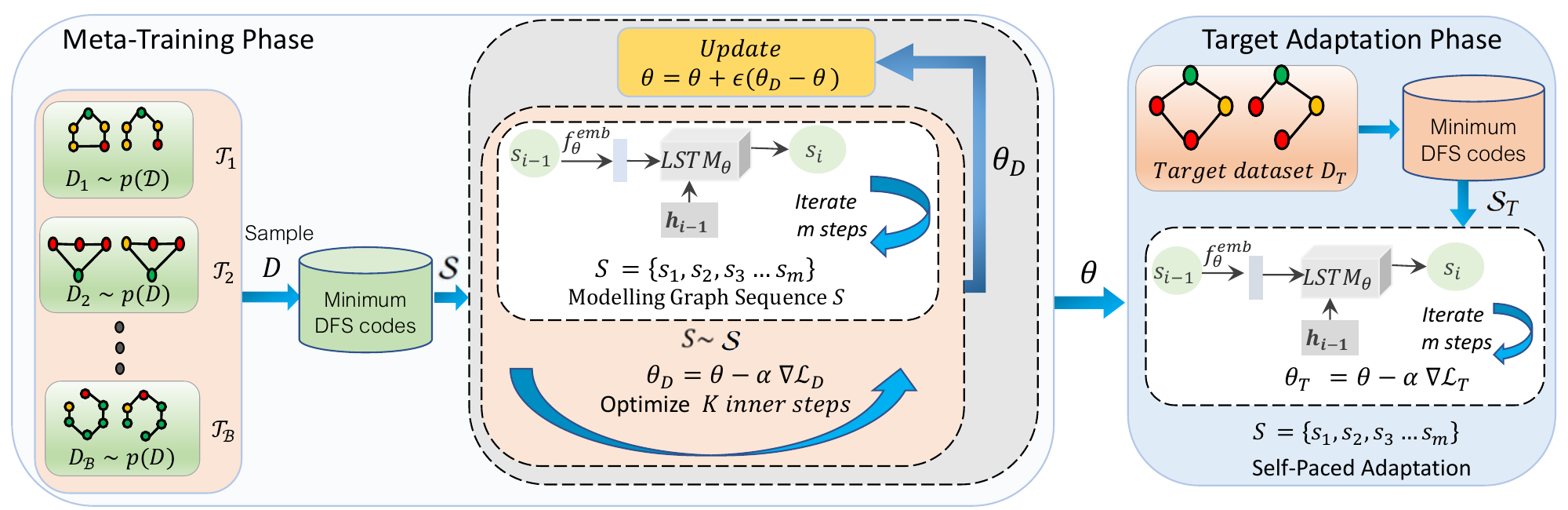}
\vspace{-0.05in}
\caption{Architecture of \namemodel}
\label{fig:model_arch}
\vspace{-0.1in}
\end{figure*}
\vspace{-0.2in}
\section{\namemodel: Proposed Methodology}
\label{sec:method}\vspace{-0.05in}

Given a set of auxiliary datasets $D_1,\cdots,D_{\mathcal{B}}$, first, \namemodel learns \textit{initial} model parameters $\theta$. $\theta$ is learned in a strategic manner such that, at inference time, when an \textit{unseen target dataset} $D_{T}$ containing a small number of graphs \rev{are} provided as input, we can fine-tune $\theta$ to new  $\theta_{{T}}$ where $p_{\theta_T}(D_T)$ best approximates the true distribution of $D_T$. Finally, to generate graphs, we sample from $p_{\theta_T}(D_T)$. Fig.~\ref{fig:framework} provides a visual summary of this approach.

The proposed approach draws inspiration from \textit{meta-learning}~\cite{finn2017model, manchanda2022generalization}. The main objective of meta-learning is to learn initial model parameters for a set of \textit{tasks} in such a way that they can be adapted to various unseen target \textit{tasks} having limited training data. In the context of our problem, each task $\mathcal{T}_i$ refers to the graph generative modeling task for dataset $D_i$ in the auxiliary dataset $\mathcal{D}$. Each task $\mathcal{T}_i$ is associated with a loss function $\mathcal{L}_{{i}}$. During meta-training, the optimal initial parameter $\theta$ is learned using $\mathcal{D}$. Then, given an \textit{unseen target task} $\mathcal{T}_{T}$ corresponding to \textit{unseen graph dataset} $D_{T}$ with an associated loss $\mathcal{L}_{{T}}$, $\theta$ is fine-tuned for $\mathcal{L}_{{T}}$ using small number of data samples of $\mathcal{T}_{T}$. When mapped to our problem of graph generative modeling, the loss function measures how well $p_{\theta_T}(D_T)$ mimics the true distribution of $D_T$.  
\vspace{-0.10in}
\subsection{Architecture Overview}
\label{sec:overview}\vspace{-0.1in}
Fig.~\ref{fig:model_arch} presents the architecture of \namemodel.  
In order to learn a generative model over labeled graphs on a dataset $D$, we first convert graphs to \textit{sequences}. This conversion allows us to leverage the rich literature on \textit{auto-regressive generative models}. Auto-regressive methods\cite{goyal2020graphgen,graphrnn} have obtained superior fidelity and high scalability on domain-agnostic graph generative modeling task. Two popular encoding schemes for encoding graphs into sequence are \textit{BFS encoding}~\cite{graphrnn} and \textit{DFS encoding}~\cite{goyal2020graphgen}. In our work, we choose \textit{DFS encoding}. This choice is motivated by the observation that  \textit{minimum DFS codes}, which is an instance of DFS encoding, provides one-to-one mapping from graphs to sequences. In contrast, in BFS encoding, the same graph may have multiple sequence representations, and may be exponential in the worst case with respect to the graph size. Consequently, one-to-one mapping is an attractive feature that our model can exploit, and as others have shown, it also improves the scalability and fidelity of graph generative modeling~\cite{goyal2020graphgen}.

Once graphs are converted into sequences via \textit{minimum DFS codes}, as shown in Fig.~\ref{fig:model_arch},  \textit{meta-learning} is conducted on the sequence representations to learn parameter set $\theta$. To model sequences, we use LSTM as shown in Fig.~\ref{fig:model_arch}. Finally, during target-adaptation phase, the target graph database $D_T$ is 
converted to the equivalent sequence representation $\mathcal{S}_T$, 
followed by fine-tuning to learn $\theta_T$. To generate graphs, we sample sequences from $p_{\theta_T}(\mathcal{S}_T)$, which are then converted to graphs. The conversion back from a sequence to its graph representation is trivial since our DFS-encoding enables one-to-one mapping. Hence, this conversion can be performed in $O(|E|)$ time, where $E$ is the set of edges. We next deep-dive into each of these individual steps.

\vspace{-0.1in}
\subsection{DFS Codes: Graph to Sequence encoding}\vspace{-0.1in}
\label{sec:dfscode}

\noindent We first formalize the concept  \textit{Graph Canonization}.\vspace{-0.05in}

\noindent
\begin{defn}[Graph Isomorphism]
\label{def:isomorphism}
Two graphs $G_i=(V_i,$ $E_i)$ and $G_j=(V_j,$ $E_j)$ are said to be \textit{isomorphic} if there exists a bijection $\phi$ such that for every vertex $v \in V_i,\; \phi(v) \in V_j$ and for every edge $e =
(u, v) \in E_i, \phi(e)=(\phi(u), \phi(v)) \in  E_j$. Furthermore, for labeled graphs to be isomorphic, in addition to above conditions, the labels of mapped nodes and edges should be same, \textit{i.e.}, $\mathbb{L}_{node}(v)=\mathbb{L}_{node}(\phi(v))$ and $\mathbb{L}_{edge}(e)=\mathbb{L}_{edge}(\phi(e))$.
\end{defn}
\begin{defn}[Graph Canonization]
Graph canonization refers to the process of converting a graph into a label such that  graphs have the same label if and only if they are isomorphic to each other. A label that satisfies this criteria is called a canonical label.
\end{defn}

\begin{figure}[t]
\vspace{-0.35in}
\centering
\subfloat[Graph G]{
\includegraphics[width=0.18\textwidth]{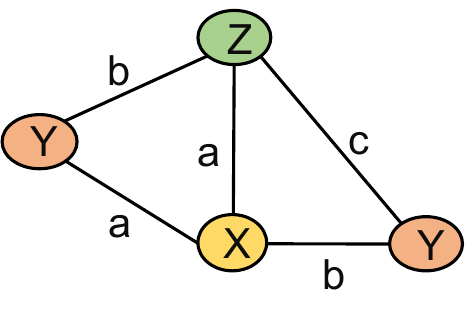}  
  }
\hspace{0.15in} 
\subfloat[DFS Code 1]{
\hspace{-0.2in}  \includegraphics[width=0.35\textwidth]{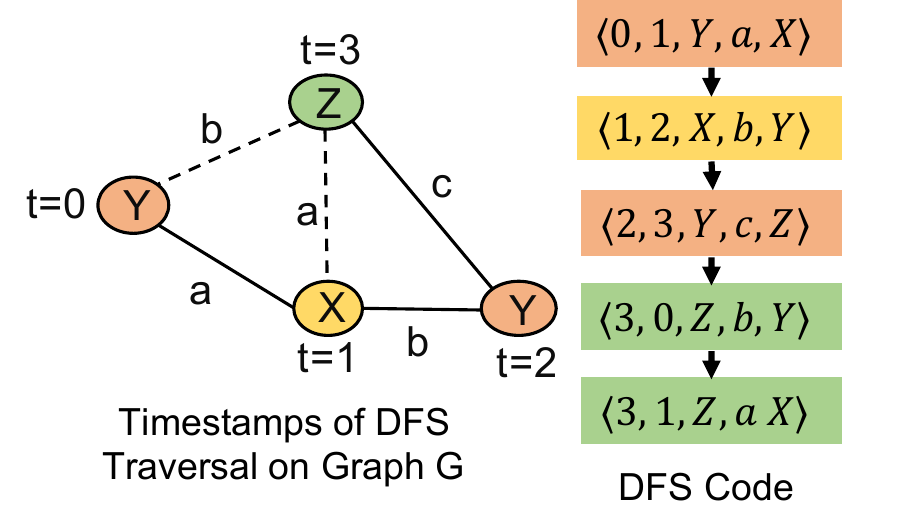}  
\label{fig:graph_to_dfs_code_1}
  }
 \subfloat[DFS Code 2]{
  \includegraphics[width=0.35\textwidth]{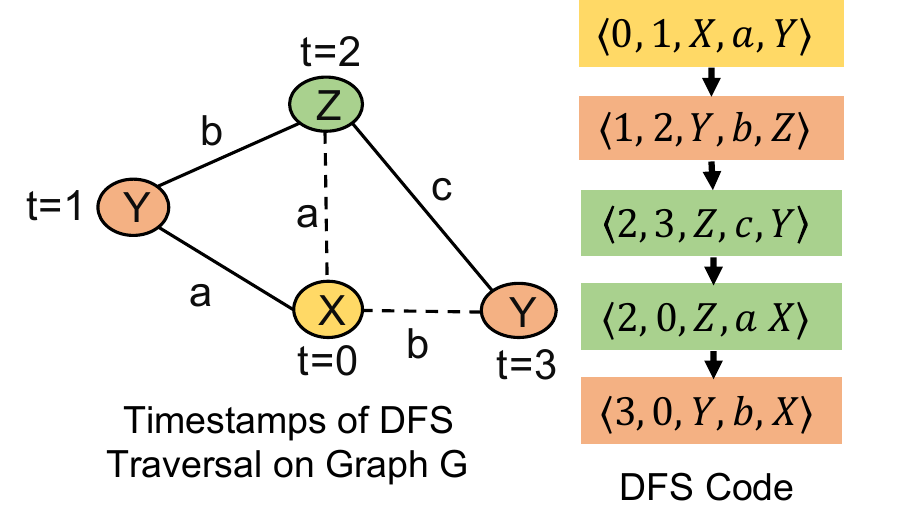}  
  \label{fig:graph_to_dfs_code_2}
  }
 \vspace{-0.05in}
 \caption{Few possible DFS codes of graph $G$. DFS code 2 is smaller than DFS code 1 since $\left\langle 0, 1, X, a, Y\right\rangle$ is less than  $\left\langle 0, 1, Y, a, X\right\rangle$}
\label{fig:graph_to_dfs_code}
\vspace{-0.3in}
\end{figure}

Now, we introduce \textit{minimum DFS codes} and how it corresponds to canonical labels of graphs. \textit{DFS code}~\cite{gspan} is a mapping function defined over a graph $G$, which encodes $G$ into a sequence of edge tuples. To construct a DFS-code from $G$, first, a depth-first search (DFS) traversal is started from an arbitrary node. During this traversal, a timestamp is assigned to each node based upon when it is discovered. The first discovered node is assigned timestamp $0$, the second discovered node is assigned $1$, and so on. Following these timestamps, each edge $(u,v)$ is assigned a tuple of five items $\left\langle t_u, t_v,L_u=\mathbb{L}_{node}(u), L_{uv}=\mathbb{L}_{edge}(uv),L_v=\mathbb{L}_{node}(v) \right\rangle$. $t_u,\;t_v$ are the discovery times of node $u$ and $v$ respectively. $L_u,\; L_v$ and $L_{uv}$ are labels of node $u$, node $v$ and edge $(u,v)$ respectively. 

A partition of edges is created based upon the DFS traversal. The first partition consists of \emph{forward edges} that are traversed by the DFS traversal. The second partition contains \emph{backward edges}, that are not traversed during the DFS traversal. For example, in Fig. ~\ref{fig:graph_to_dfs_code_1},~\ref{fig:graph_to_dfs_code_2}, the edges depicted by solid lines depict the forward edges, and the one's which are dashed depict backward edges. A \textit{total ordering} is imposed on these edges following the rules described in \textsc{gSpan} \cite{gspan} to obtain the DFS code of a graph. Specifically, for ordering forward edges, the process is straight forward. Forward edges are ordered based upon their discovery time in the DFS traversal. For backward edges, the ordering is derived based upon the following rules:
    
\begin{itemize}
    \item Backward edge $(u,s)$ must appear before all forward edges of the form $(u,t)$.
    \item Backward edge $(u,s)$ must appear after the forward edges of the form $(t,u)$, i.e the first forward edge which points to $u$.
    \item For backward edges of the form $(u,s)$ and $(u,s')$  originating from the same source $u$,  $(u,s)$ is ordered before $(u,s')$ if $t_s < t_s'$.
\end{itemize}
Fig.~\ref{fig:graph_to_dfs_code} shows  examples of two DFS codes of a graph based on two DFS traversals. For more details on DFS code, we refer to \textsc{GSpan}\cite{gspan}.

 \textbf{Minimum DFS codes:} As shown in Fig.\ref{fig:graph_to_dfs_code}, a graph can have multiple DFS codes. We choose the lexicographically smallest DFS code among all DFS codes as the \textit{minimum} DFS code. It has been shown that there exists a \textit{bijection} between a graph and its minimum DFS code~\cite{gspan}. Hence, minimum DFS codes are canonical labels. Using minimum DFS codes, we encode each graph $G=(V,E)$ in dataset $D$ as a sequence of $m$ edge tuples $S=(s_1,\ldots,s_m)$ where $m=|E|$  and each $s_i$ is an edge tuple of the form $\left\langle t_u, t_v,L_u, L_{uv},L_v\right\rangle$. We use the notation $\mathcal{F}(G)=S$ to denote the minimum DFS code $S$ of graph $G$. 
Applying $\mathcal{F}$ on all graphs of dataset $D$, we obtain a collection of edge tuple sequences $\mathcal{S}=\left\{\mathcal{F}(G) \mid \forall G \in D\right\}$
for all graphs in dataset $D$. 

\textbf{Computation Complexity:} We note that computing the minimum DFS code of a graph is equivalent to performing graph isomorphism tests. In the literature, no polynomial time algorithm exists for detecting graph isomorphism. Fortunately, for labeled graphs, it has been shown that minimum DFS codes can be computed very efficiently\cite{gspan,goyal2020graphgen}.

\vspace{-0.15in}
\subsection{Modeling Graph Sequences}\vspace{-0.1in}
   Minimum DFS codes are of sequential nature. We model each sequence $S{=}(s_1,\cdots, s_m)$ using an \textit{auto-regressive} model\cite{goyal2020graphgen} as follows:  \vspace{-0.07in}
\begin{equation}\vspace{-0.04in}
    \label{eq:conditionaldecomp}
    p(S) = p(s_0)\prod_{i = 1}^{m + 1}p(s_i|s_0,\cdots,s_{i-1})
\end{equation}
where $m{=}{{\mid}{E}{\mid}}$ is the number of edges, $s_0$ is a \emph{start-of-sequence} SOS token and $s_{m+1}$ is \emph{end-of-sequence} EOS token to allow variable length sequences. To learn the parameters for these sequential conditional distributions, we use \textit{Recurrent Neural Networks}. Specifically, we use LSTM~\cite{10.1162/neco.1997.9.8.1735}, which efficiently models long-range dependencies. Formally,
\begin{align}\vspace{-0.05in}
    \label{eq:lstm}
        \ch_i &= LSTM^{hidden}_{\theta}\left(\ch_0,\left(f^{emb}_{\theta}(s_0) \ldots f^{emb}_{\theta}(s_{i-1})\right)\right)  &= LSTM^{hidden}_{\theta}\left(\ch_{i-1},f^{emb}_{\theta}\left(s_{i-1}\right)\right)
\end{align}
where $LSTM_{\theta}$ is a function representing an LSTM cell. $f^{emb}_{\theta}$ is an embedding function that takes \textit{one-hot encoding} of $s_{i-1}$ as input and produces a $d$-dimensional compressed vector. $\ch_0$ is initialized to $\boldsymbol{0}$. Finally, assuming that $s_i.t_u$, $s_i.t_v$, $s_i.L_u$, $s_i.L_{uv}$, $s_i.L_v$ are independent\footnote{\rev{In Sec. ~\ref{app:cond_edge_tuple} in Appendix, we study the impact of  predicting the components of edge tuple conditionally instead of independently.} } given $\ch_i$, we predict $s_i=\langle t_u, t_v,L_u, L_{uv}, L_v\rangle$ as follows. \vspace{-0.05in}
\begin{align}
    \label{eq:s_i}
    \widetilde{s_i} = \left\langle f^{t_u}_{\theta}(\ch_i),f^{t_v}_{\theta}(\ch_i),f^{L_u}_{\theta}(\ch_i),f^{L_{uv}}_{\theta}(\ch_i),f^{L_v}_{\theta}(\ch_i)\right\rangle
\end{align}\vspace{-0.01in}
where each $f_{\theta}$ is a function  representing a fully connected \textit{Multi-layered Perceptron (MLP)}. Note that every function in this discussion is parameterized by $\theta$ (indicated by the subscript). Finally, we define the loss $\mathcal{L}_D$ specific to sequence (graph) generation task $\mathcal{T}_D$ on dataset $D$ as follows: \vspace{-0.05in}
\begin{align}
    \mathcal{L}_S = -\sum_{i=1}^{m+1}\sum_{c}\left(s_{i}[c] \log \widetilde{s}_{i}[c]+\left(1-s_{i}[c]\right) \log \left(1-\widetilde{s}_{i}[c]\right)\right) 
    & \quad \text{,}  \quad 
    \mathcal{L}_D = \sum_{S \in \mathcal{S}} \mathcal{L}_S 
    \label{eq:Ld-Ls}
\end{align}
\looseness=-1
Here $c$ is the component index of one-hot vector $s_i$ and predicted vector $\widetilde{s}_i$. $\mathcal{S}$ is the collection of graph sequences $S$ derived by encoding every graph $G{\in} D$ using minimum DFS coding function $\mathcal{F}(G)$.
 
\vspace{-0.05in}
\subsection{Meta-Learning for Labeled Graph Generative Modeling under limited availabiltiy of data}\vspace{-0.1in}
Up until now, we have defined parameters $\theta$ of graph generative model $p_{\theta}(D)$. As motivated, we want to find an initialization of $\theta$ such that it can quickly learn to generate graphs from unseen dataset $D_T$ having small number of training graph samples. Specifically, we train $\theta$ on graphs from auxiliary datasets to learn initial parameters. To do this, we build upon the \textsc{Reptile} framework~\cite{nichol2018reptile}. \textsc{Reptile} is a\textit{ first-order} meta-learning algorithm, wherein it uses first-order gradients to learn $\theta$, and  is therefore computationally and memory efficient. \namemodel, using \textsc{Reptile}, extracts the \textit{meta-knowledge} to obtain an effective initialization and an ability to adapt to the target dataset using limited fine-tuning samples. More concretely, \namemodel optimizes the below objective function in order to learn good initialization of $\theta$: 
\begin{equation}
    \label{eq:meta-pb}
    \min_\theta \, 
    \mathbb{E}_{{\mathcal{L}_{D \sim \mathcal{D}}}} \left[\mathcal{L}_{D} (\theta_{D}^{K})\right],\vspace{-0.1in}
\end{equation}\vspace{0in}
where $\theta_{D}^{K}$ are the updated parameters after $K$ gradient updates of $\theta$ from dataset 
${D}$ as follows:
\begin{align}
\label{eq:gradient_descent_D}
 \theta^0 = \theta  & \quad \text{and}\quad \quad
 \theta_{D}^i=\theta_{D}^{i-1} -\alpha \nabla_{\theta^{i-1}}\mathcal{L}_D \;\;\; \forall i \in [1 \ldots K]
\end{align}
Here hyper-parameter $\alpha$ controls the meta-learning rate.
\noindent
Finally  using the $K$ step updated parameters     $\theta^K_D$, we optimize Eq.~\ref{eq:meta-pb} as follows:\vspace{-0.1in}
\begin{equation}
    \label{eq:meta_gradient}
    \theta = \theta + \epsilon\left( \theta^K_D - \theta \right)
\end{equation}
where $\epsilon$ and $K$ are hyper-parameters of \namemodel. Eq.~\ref{eq:meta_gradient} updates the value of the meta-parameters $\theta$ using a weighted combination of $\theta$ and \textit{K-step} fine-tuned parameter $\theta^K_D$ for dataset $D$. The parameter  $\epsilon$ can be considered as a step-size in the direction of the gradient $\theta^K_D - \theta $.  We iterate over $D{\sim}\mathcal{D}$ by computing Eq.~\ref{eq:gradient_descent_D} for different tasks and then using it for optimizing Eq.~\ref{eq:meta-pb}. Algorithm ~\ref{alg:meta_training} in App. describes the pseudocode of meta-training procedure of \namemodel.

\subsection{Adaptation to Target Dataset}
\vspace{-0.1in}
Once \namemodel is meta-trained on diverse graph datasets, our next goal is to adapt the learned model parameters to the target dataset ${D_T}$. 
Essentially, first we initialize the target model parameters to the value of the meta-trained model: \vspace{-0.15in}
\begin{align}
   \theta_{T} = \theta \tag{Initialization}
%
\end{align}
Towards our goal to optimize parameters on the target dataset, a simple approach is to update the parameters of the model by applying multiple gradient updates using samples from target dataset $D_T$ with its associated loss  as follows:\vspace{-0.12222222in}
\begin{align}
\vspace{-0.1in}  \theta_{T} = \;& \theta_{T} -\alpha  \nabla_{\theta_{T}}\mathcal{L}_{T}  \tag{gradient updates} 
\end{align}
The above equation assumes, for every gradient update, the training data is sampled in a random fashion from the target dataset. However, recent studies have discovered that gradually increasing the complexity of training instances results in better learning and faster convergence~\cite{zaremba2015learning}. Motivated by this result, we adopt \textit{self-paced learning}~\cite{NIPS2010_e57c6b95} in the fine-tuning phase of \namemodel. Towards this end, we modify the loss $\mathcal{L}_{T}$ associated with the target dataset in a way that the model is presented with training samples of gradually increasing difficulty. Moreover, the training curriculum is dynamically determined by the model itself based upon its \textit{perception} of the difficulty of a sample. Specifically, recall from Eq.~\ref{eq:Ld-Ls} $\mathcal{L}_D = \sum_{S \in \mathcal{S}} \mathcal{L}_S $
where $\mathcal{S}$ is the collection of graph sequences of $G\in D$. For self-paced learning, we modify $\mathcal{L}_{T}$ as follows: \vspace{-0.05in}
\begin{equation}
\begin{aligned}
\label{eq:self_pace_loss}
    \mathcal{L}_{T} &= \sum_{i=1}^{\mid \mathcal{S}_T \mid}\beta_i\mathcal{L}_{S_i} -\lambda\sum_{i=1}^{\mid \mathcal{S}_T \mid} \beta_i & \quad \quad
    \beta_i \in & \{0,1\}\;\; \forall i \in [1 \ldots {\mid} \mathcal{S}_T{\mid}]
\end{aligned}
\end{equation}
where $\mathcal{S}_T{=}\{\mathcal{F}(G) \mid \forall G \in D_T\}$, $S_i{\in}\mathcal{S}_T$, and ${{\mid}{\mathcal{S}_T}{\mid}}$ is the number of graphs in $D_T$. $\lambda$ is an evolving parameter that essentially controls the pace of learning. Specifically in our graph generative modeling setting, we solve this via an iterative approach~\cite{NIPS2010_e57c6b95}. Before every gradient update as described earlier, we first calculate the value of $\beta_i$'s as follows:\vspace{-0.02in}
\begin{equation}
    \beta_i= 
\begin{cases}
    1 & if \;\; \mathcal{L}_{S_i} < \lambda \\
    0 & else
\end{cases}
\end{equation} 
The value of $\beta_i$ indicates  whether the $i^{th}$ training sample will be used or not in the loss computation in Eq.~\ref{eq:self_pace_loss} . We substitute these values in Eq.~\ref{eq:self_pace_loss} and update the parameters $\theta_{T}$. This process repeats until convergence. The value of $\lambda$, is increased periodically by a growth factor $\gamma$ to gradually allow hard samples to be a part of the loss computation during the course of training. Algorithm  ~\ref{alg:fine_tuning} in App. describes the pseudocode of the fine-tuning procedure of \namemodel.
\vspace{-0.1in}

\subsection{Graph Generation}
\label{app:graphgeneration}
\vspace{-0.1in}
After fine tuning $p_{\theta}(D)$ on target dataset $D_T$, we obtain $p_{\theta_T}(D_T)$. We sample graphs from this distribution as follows. First, we pass the initial hidden state $\ch_0{=}\boldsymbol{0}$ to $LSTM_{\theta_{T}}$ along with the SOS symbol. At each step $i$, we sample $s_i$ from the updated hidden state $\ch_i$ as follows-
\begin{align}
\nonumber
   \label{eq:generation}
    s_i.t_u & \sim \; Multinomial(f^{t_u}_{\theta_{T}}(\ch_i)) & \quad 
    s_i.t_v & \sim \; Multinomial(f^{t_v}_{\theta_{T}}(\ch_i)) \\
    s_i.L_u & \sim \; Multinomial(f^{L_u}_{\theta_{T}}(\ch_i)) & \quad 
    s_i.L_{uv} & \sim \; Multinomial(f^{L_{uv}}_{\theta_{T}}(\ch_i)) \\
    s_i.L_v & \sim \; Multinomial(f^{L_v}_{\theta_{T}}(\ch_i))
\end{align}
The process is repeated until the EOS symbol is sampled for any of the five components in the sampled tuple. Finally, this sampled sequence, representing the DFS code, is converted back to graph. Algorithm~\ref{alg:graph_generation} in App. presents the pseudocode of the graph generation phase.

\vspace{-0.1in}\section{Experiments}
\label{sec:exp}\vspace{-0.1in}
We benchmark \namemodel against state-of-the-art algorithms for graph generation and establish that:
\begin{itemize}
    \item \textbf{Higher fidelity:} \namemodel generates graphs of higher fidelity than the state-of-the-art methods.
    
    \item \textbf{Sample-efficient:} Attributed to its capability to learn with a limited amount of data,  \namemodel better preserves graph properties compared to existing methods even when the number of fine-tuning samples used by \namemodel are relatively less compared to other methods.
\end{itemize}

The code to reproduce the experiments can be found at \url{https://github.com/idea-iitd/GShot}.

    
\subsection {Experimental setup}

\textbf{Datasets}:\label{sec:datasets}\label{sec:dataset} Since our focus is on \textit{domain-agnostic labeled}  graph generative modeling, we show the effectiveness of our proposed approach using datasets from diverse domains. Moreover, in our experiments, we use target datasets having significantly low volumes of available graphs in comparison to other works in literature \cite{goyal2020graphgen,graphrnn,gram,gran}. Table ~\ref{tab:datasets} in App. ~\ref{app:dataset-desc} summaries the different datasets. Further details on semantics of the datasets are also present in App.~\ref{app:dataset-desc}.

\textbf{Train-test splits:} We briefly describe our datasets' train-test split.

\vspace{-0.05in}
\begin{itemize}
        \item \textbf{Biological Domain:} 
         Each enzyme in the Enzyme dataset\cite{bogwartenzymes} belongs to one of six classes, namely EC1, EC2, EC3, EC4, EC5, EC6.
        We treat enzymes in EC1, EC2, EC4, EC5, EC6 as auxiliary datasets and EC3 as our target dataset, which consists of $100$ enzymes. \footnote{\rev{In Sec.~\ref{app:sim_tar_aux} in Appendix we find similarity of target datasets with auxiliary datasets and show impact of performance of \namemodel using different auxiliary datasets having different similarity.} }
    
    \item \textbf{Chemical Domain:}
    We use anti-cancer screen datasets  \textit{Yeast}, \textit{Breast}, and \textit{Lung} as auxiliary datasets for meta-training and use the two smallest chemical datasets of \textit{AIDS-CA} and \textit{Leukemia-Active}  as our target set. 
    

    \item \textbf{Physics Domain:} 
    We meta-train \namemodel on auxiliary datasets consisting of \textit{four} and \textit{six} particle spring systems and then fine-tune on graphs containing \textit{five} particles.
\end{itemize}

\looseness=-1
\textbf{Baselines:}We benchmark the performance of \namemodel against the state-of-the-art techniques for \textit{domain-agnostic, labeled} graph generative modeling, namely \graphGen~\cite{goyal2020graphgen} and \graphRNN~\cite{graphrnn}. We do not include \textsc{GRAN}\cite{gran} as a baseline since it cannot generate labeled graphs. For \graphGen, we used the code shared by authors. While, in theory, \graphRNN  supports labeled graphs, the code shared by the authors do not. Hence, we extend the author's code as outlined in the supplementary section of \graphRNN\cite{graphrnn}. Both \graphGen and \graphRNN are  trained \textit{only} on the target dataset and we compare the quality of the generated graphs with that of \namemodel. This comparison allows us to evaluate how efficient the knowledge transfer of \namemodel is as opposed to relying only on the target dataset. 

In addition, we use a third \textit{pre-training} baseline introduced by us, which we will refer to as \textsc{PreTrain+FT}. In this baseline, we first pre-train \graphGen on the same auxiliary datasets used by \namemodel for meta-training. Then, we fine-tune it on the target dataset. This baseline allows us to systematically understand the impact of meta-learning against generic generative modeling. We do not consider pre-training on \graphRNN, since \graphGen has been shown to be superior on the labeled graph generative modeling task~\cite{goyal2020graphgen}, which is also reflected in our experiments that follows.

\textbf{Evaluation setup:}
\label{sec:eval:main}
During meta-training of \namemodel, we use ${\approx}50\%$ data for training and the same for validation. During fine-tuning to a new graph dataset, unless specifically mentioned, we use the default split among training, validation, and test as ${\approx}40\%$, ${\approx}30\%$, and ${\approx}30\%$ respectively.  
Unless specified otherwise, for training a model from scratch directly on the target dataset or fine-tuning a model on a target dataset, we use the same number of training samples of the target dataset. For each target dataset, this information is present in the  \textit{\#Target Training samples} column of Table ~\ref{tab:quality}. The  system configuration and parameter details can be found in  App.~\ref{sec:app:exp}. 
 

\textbf{Evaluation Metrics:}\label{sec:metrics} The performance of a graph generative model is satisfactory {\bf (1)} if it generates graphs with similar properties as the source graphs, {\bf (2)} but without duplicating the source graphs themselves. To quantify these, we divide our metrics into two categories.\vspace{-0.1in}
\label{sec:eval_metrics}

\noindent
\begin{itemize}
    \item \textbf{Fidelity:} To quantify the preservation of graph properties,  we compare the distributions of graph statistics between the ground truth graphs and the generated graphs using the following metrics.
\begin{itemize}
    \item \textbf{Structural metrics:} To quantify the preservation of original graph properties, we use the structural metrics used by GraphRNN and GraphGen: {\bf (1)} \emph{node degree distribution} (Degree), {\bf (2)} \emph{clustering coefficient distribution of nodes} (Clustering), and {\bf (3)} \emph{orbit count distribution} (Orbit)~\cite{hovcevar2014combinatorial}, which measures the number of orbits with $4$ nodes. This metric captures the higher-level motifs that are shared between generated and test graphs. We utilize \emph{Maximum Mean Discrepancy (MMD)}~\cite{metric:mmd} to compute the distance between two distributions. Further, to compare the sizes of the generated graphs against the ground truth, we measure {\bf (4)} \emph{Average node count} and {\bf (5)} \emph{Average edge count}.

    \item \textbf{Labeled Graph Metrics:} Our work is geared towards labeled graph generation. Hence, it is important to assess   whether a generative model captures the label distribution well. Towards that end, we compare the distribution of {\bf (1)} Node Labels, {\bf (2)} Edge Labels, and {\bf (3)} the \textit{joint distribution} of node labels and degree in the ground truth and generated graphs. We again use MMD to quantify the distance from the ground truth.
    
    \item\textbf{Topological Similarity:} Finally, in order to capture topological similarity of generated graphs with the ground truth graphs, we use \emph{Neighbourhood Sub-graph Pairwise Distance Kernel (NSPDK)}~\cite{nspdk}. NSPDK provides  the benefit of incorporating both node and edge labels along with the structure of the graph. Specifically, NSPDK measures the distance between two graphs by matching pairs of subgraphs with different radii and distances. The lower the MMD score for NSPDK, the more aligned are the two graph distributions.
    \end{itemize}
   
    \item \textbf{Duplication and Uniqueness:} A model that generates graphs with high fidelity might not be useful in practice unless it is also capable of generating graphs that are not seen in the training data.
    In order to capture this requirement, we utilize the below metrics introduced by \graphGen \cite{goyal2020graphgen}: {\bf (1)} \emph{Novelty} measures the percentage of generated graphs that are not subgraphs of the set of the training graphs. Additionally, we compute {\bf (2)} \emph{Uniqueness}, which captures the diversity of the set of generated graphs. In order to quantify uniqueness, we remove the generated graphs that are subgraph isomorphic to any of the other generated graphs. This is different from the novelty metric as here we focus only on the generated graphs. A model that generates $100$ graphs and out of which $90$ are subgraph isomorphic to any of the other generated graphs has uniqueness${=}10\%$.
\end{itemize}


In order to quantify the quality of a particular metric, we generate multiple graphs for each target dataset and compare them against the available ground truth target graphs. Details of number of graphs generated for each dataset is present in   App.~\ref{sec:app:graphs_gen}.

\renewcommand{\tabcolsep}{3pt}
\begin{table*}[t]
\vspace{-0.20in}
\caption{Summary of performance by \namemodel, \graphGen, \graphRNN, and \textsc{PreTrain+FT} baseline on different datasets on multiple metrics. Values less than $10^{-3}$ are approximated to 0. The best-performing model for each dataset is highlighted in bold.}
\vspace{-0.1in}
\label{tab:quality}\hspace{-0.1in}
\small
\scalebox{0.8}{
{\scriptsize
\begin{tabular}{p{5em}|p{3.8em}|p{3.8em}|c|ccc|c|cc|ccc|cc} 
    \textbf{Auxiliary datasets} &\textbf{Target dataset} & \#\textbf{Target Training Samples} & \textbf{Model} & \textbf{Deg.} & \textbf{Clus.} & \textbf{Orbit} & \textbf{NSPDK}  & \makecell{\textbf{Avg} \# \textbf{Nodes} \\ \textbf{(Gen/Gold)}} & \makecell{\textbf{Avg} \# \textbf{Edges} \\ \textbf{(Gen/Gold)}} & \makecell{\textbf{Node} \\ \textbf{Label}} & \makecell{\textbf{Edge} \\ \textbf{Label}} & \makecell{\textbf{Joint} \\ \textbf{Node} \\ \textbf{Label} \\ \& \textbf{Degree}} & \textbf{Novelty} & \textbf{Uniqueness}  \\
    
    \midrule
    \multirow{4}{2em}{Enzyme: EC1,EC2, EC4,EC5,  EC6} &
    \multirow{4}{4em}{\\Enzyme: EC3} &\multirow{4}{4em}{\\50}&\textsc{\graphGen}  & 0.90 & 0.58 & 0.127 &0.266 & 17.51/26.90 & 22.22/52.85 & 0.015 & x & 0.714 & \textbf{100\%} & \textbf{100}\% \\
    
   & & & \graphRNN & \textbf{0.30} & 0.73 & 0.13 & 0.214 & 20.51/26.90 & 36.23/52.85 & 0.019 & x & 0.696 & \textbf{100\%} & \textbf{100\%}  \\

   & && \textsc{PreTrain+FT} & 0.72 & 0.63 & 0.053 & 0.18 & 23.73/26.90 & 33.2/52.85 & 0.0095 & x & 0.619 & $99\%$ & $99\%$  \\

 &&& \namemodel & 0.45 & \textbf{0.47} & \textbf{0.025} & \textbf{0.16} & \textbf{24.5/26.90}  & \textbf{37.69/52.85} &\textbf{0.004}  & x & \textbf{0.457} & \textbf{100\%} & \textbf{100}\%\\
 
    \hline   
    \multirow{8}{4em}{\\Yeast, Breast, Lung} &
    \multirow{4}{4em}{\\AIDS-CA} & \multirow{4}{4em}{\\150}& \textsc{GraphGen} & 0.026 &  $0.016$ & $0.003$ & 0.127 & 17.51/37.14 & 17.62/39.60& 0.05 & $0.001$ & 0.20 & $98\%$ & $97\%$  \\

&    && GraphRNN & 0.15 & 0.47 & 0.045 & 0.14 & \textbf{30.5/37.14} & \textbf{40.19/39.60} & 0.193 & 0.005 & 0.836 & $86\%$ & $45\%$  \\

     &&& \textsc{PreTrain+FT} & 0.021 & 0.004 & $\approx 0$ & 0.11 & 24.1/37.14 & 25.22/39.60 & 0.013 & \textbf{$\approx$ 0} & 0.173 & $\textbf{99\%}$ & $\textbf{99\%}$ \\
     
     &&& \namemodel & \textbf{0.017} & \textbf{0.0015} & \textbf{$\approx$ 0} & \textbf{0.08} & 26.5/37.14 & 27.1/39.60 & \textbf{0.011} & \textbf{$\approx$ 0} & \textbf{0.14} & \textbf{99\%} & \textbf{99\%}  \\

    \cline{2-15}
     &
    \multirow{4}{4em}{Leukemia-Active} &\multirow{4}{4em}{\\500}&\textsc{GraphGen} &0.06 & $0.019$ & $\approx 0$ &0.17 &40.02/47.71 &42.23/50.37 & 0.02 & \textbf{$\approx$ 0} & 0.99 & \textbf{100}\% & \textbf{100}\% \\

   & && GraphRNN & 0.06 & 0.554 & 0.032 & 0.34 & 7.17/47.71 & 7.51/50.37 & 0.39 & 0.017 & 0.83 & \textbf{100}\% & \textbf{100}\%  \\

    &&& \textsc{PreTrain+FT} & 0.039 & 0.0064 & $\approx$ 0& 0.116 & \textbf{43.08/47.71} & \textbf{45.5/50.37} & 0.09 & \textbf{$\approx$ 0} & 0.79 & $98\%$ & $98\%$  \\
    
     &&& \namemodel & \textbf{0.0069} & \textbf{$\approx$ 0} & \textbf{$\approx$ 0} & \textbf{0.032 }& 42.35/47.71  & 44.33/50.37 & \textbf{0.0011}  & \textbf{$\approx$ 0} & \textbf{0.24} & \textbf{100}\% & \textbf{100}\% \\

\hline
\multirow{4}{4em}{\\\{4, 6\} body Spring} &
    \multirow{4}{4em}{\\5-body Spring} &\multirow{4}{4em}{\\500}&\textsc{GraphGen} & \textbf{0.004} & \textbf{0.015} & \textbf{$\approx$ 0} & \textbf{0.016} & \textbf{4.98/5} & \textbf{5.49/5.64} & 0.012 & x  & \textbf{0.011} & 33\% & 13\%  \\

    &&& GraphRNN & 0.018 & 0.012 & \textbf{$\approx$ 0} &0.029 & 4.71/5 & 5.03/5.64 & 0.044 & x  & 0.017 & \textbf{87\%} &\textbf{ 55\%}  \\

    &&& \textsc{PreTrain+FT} & 0.021 & 0.047 & 0.0025 & 0.017 &\textbf{ 4.98/5 }& 5.19/5.64 &\textbf{ 0.011 }& x & 0.012 & $70\%$ & $49\%$ \\
   
   &&& \namemodel & 0.008 & 0.035 & \textbf{$\approx$ 0} & \textbf{0.016} & \textbf{4.98/5 }& 5.38/5.64 & 0.016 & x  & 0.012 & 64\% & 41\%  \\
    \bottomrule
\end{tabular}}
}
\vspace{-0.20in}
\end{table*}
\vspace{-0.1in}

\subsection{Quality}\vspace{-0.1in}
\label{sec:quality}
\noindent \textbf{Fidelity:} Table~\ref{tab:quality} shows the performance of all the models across different datasets. We observe that, in most cases, \namemodel obtains lower MMD scores compared to baselines. In terms of the global level graph metric \textit{NSPDK}, \namemodel achieves a significant improvement even against the best performing baselines. For instance in Leukemia-Active, for \textit{NSPDK}, \namemodel obtains MMD value of $0.032$ against a significantly higher value of $0.116$ obtained by \textsc{PreTrain+FT}. With respect to labeled graph metrics, we observe that \namemodel improves over state-of-the-art techniques by achieving more than $50\%$ lower MMD value in multiple cases. Further, \namemodel also outperforms existing techniques in the \textit{Joint Node-Label and Degree} metric signifying its ability to better jointly model the graph structure and labels.  \rev{While \namemodel achieves high quality on most fidelity metrics, however on the node/edge count metric when the number of fine-tuning graphs becomes very low(Eg:- Enzyme EC3 = 50  and AIDS-CA = 150), its  performance deteriorates for this metric. We would also like to highlight that modeling the edge tuple conditionally in \namemodel(see App ~\ref{app:cond_edge_tuple}), improves this metric to a certain extent on all datasets. Among all baselines only \graphRNN's performance on AIDS-CA for avg node/edge count is  superior to \namemodel, however it performs worser on other fidelity metrics .} Having said that, the superior performance of \namemodel establishes the efficacy of the meta-training procedure to learn an effective set of initial model parameters, which adapts well to  low-data regimes.

\noindent \textbf{Uniqueness and novelty:}
In addition to obtaining better fidelity in most cases, \namemodel also achieves a higher or similar score compared to baselines on the \textit{Novelty} and \textit{Uniqueness} aspect. On AIDS-CA dataset, we obtain an improvement of ${\approx}1{-}2\%$ in the uniqueness and novelty metrics against \graphGen while also  achieving better fidelity scores. Additionally, for AIDS-CA, \graphRNN's performance in terms of both fidelity as well as diversity is significantly inferior to other methods. In the case of the $5$-body spring dataset, although \namemodel does not perform the best in terms of fidelity scores, still its uniqueness and novelty scores are significantly higher than \graphGen, which achieves a $33\%$ novelty score and $13\%$ uniqueness score. This indicates that \graphGen mostly generated duplicated graphs. \rev{In context of Physics dataset, we highlight that the target graphs are extremely small i.e having only 5 nodes. It is more challenging to generate unique and diverse graphs in this context when compared to datasets belonging to other category such as chemical/biological where graphs are large. We observe that only GraphRNN is able to generate graphs which are more novel and unique, however, its performance is significantly poor on all fidelity metrics.}  Overall, we observe that an efficient parameter initialization obtained by \namemodel also helps in improving the diversity of generated graphs while generating graphs with high  fidelity. 
\begin{figure}[h]
\vspace{-0.1in}
\centering
\includegraphics[scale=0.22]{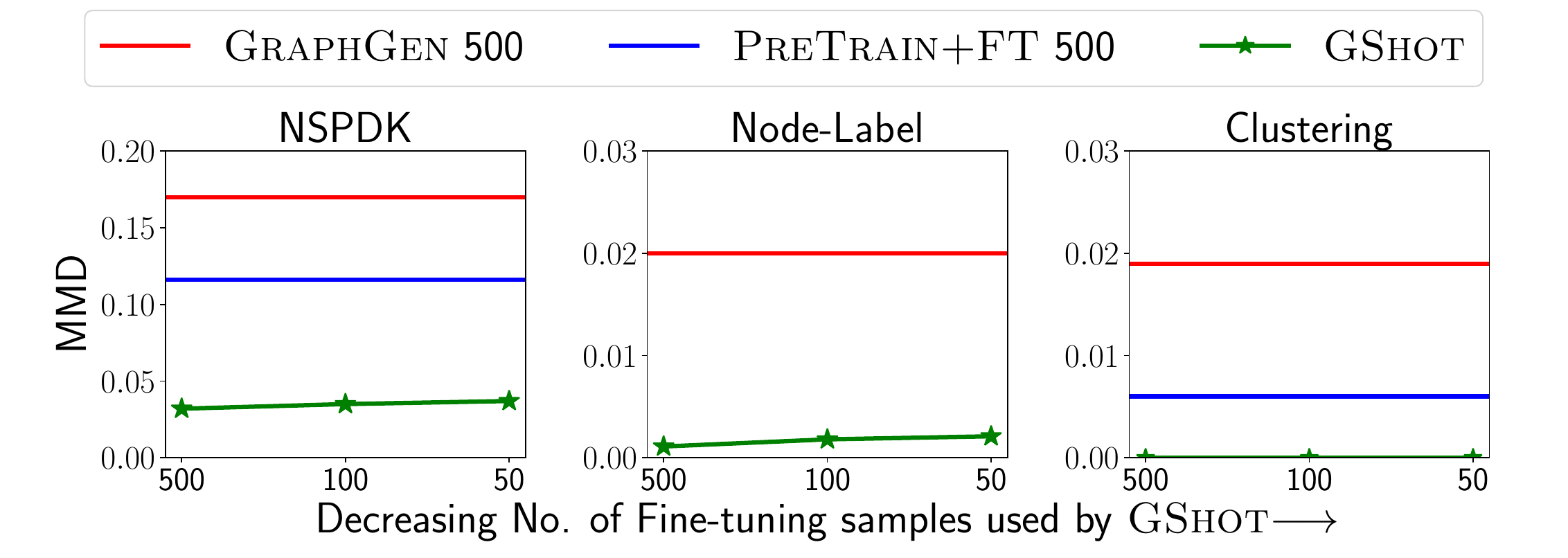}  
 \caption{\label{fig:lessdatasample} The variation in MMD scores on different metrics when the number of fine-tuning samples for \namemodel \rev{is} reduced from $500$ to $50$ on the Leukemia-Active dataset. Here the suffix of $500$ after \graphGen and \textsc{PreTrain+FT} depicts that the number of training samples used from the target dataset for these baselines is $500$. Note that for \textsc{PreTrain+FT} 500, the value of the Node-Label metric ($0.09$) was skipped in the diagram to improve readability. }\vspace{-0.1in}
\vspace{-0.05in}
\end{figure}

\noindent\textbf{Robustness to number of fine-tuning samples:} We also evaluate \namemodel's robustness to different sizes of the same fine-tuning dataset. Towards this end, we choose Leukemia-Active dataset as our target dataset since due to slightly higher availability of fine-tuning data, there is a reasonable scope to down-sample the  fine-tuning data in order to understand its impact on performance. We vary the number of fine-tuning samples available to \namemodel from $500$ to $50$. However, we keep the number of fine-tuning samples for the baselines to the maximum value, i.e., $500$. In Fig.~\ref{fig:lessdatasample}, we observe that \namemodel, while using less number of samples from the target dataset, still obtains lower MMD scores on different metrics in comparison to \graphGen and the \textsc{PreTrain+FT} model that used $500$ samples from the target dataset. Further, the MMD scores for \namemodel increase only slightly  when the number of fine-tuning samples is reduced from $500$ to $50$. This is a direct consequence of our model's ability to adapt with a small number of training samples. Further, we would like to highlight that the novelty and uniqueness metrics did not show any observable change in this experiment.

\noindent \textbf{Ablation study:} We study the improvement obtained by using \textit{self-paced fine-tuning} in  \namemodel over {\textit{vanilla fine-tuning}} in Appendix~\ref{app:self_pace}. In Appendix ~\ref{var:aux:dataset}, we study the impact of the choice of auxiliary datasets on the performance of graph generative modeling under data scarcity.
\vspace{-0.1in}
\section{Conclusion} 
\label{sec:conc}\vspace{-0.05in}
 Research on deep graph generative modeling has progressed significantly in several directions such as scalability to large graphs,  domain agnostic modeling, handling node and edge labels, etc. However, the problem of learning to generate graphs in low-data regimes remained unexplored. In this work, we propose the paradigm of \textit{ domain-agnostic, labeled graph generative modeling under data scarcity}. Our proposed architecture \namemodel learns to transfer meta-knowledge from auxiliary graph datasets to a target dataset. Utilizing these prior experiences, \namemodel quickly adapts to an unseen graph dataset through \textit{self-paced fine-tuning}. \namemodel is effective in learning graph distributions on datasets with small number of available training samples. Extensive evaluation on real graph datasets demonstrate that graphs generated by \namemodel preserve graph structural properties significantly better than the state-of-the-art approaches. Although, our proposed method outperforms existing state-of-the-art methods, however, while generating these molecules, it does not take into account their molecular/chemical properties etc. In future, we would like to work on capturing these aspects.\vspace{0.1in}
\vspace{-0.1in}\section{Acknowledgement}\vspace{-0.05in}
Srikanta Bedathur was partially supported by a DS Chair of AI fellowship. Sayan Ranu acknowledges the Nick McKeown chair position endowment. Sahil Manchanda was partially supported by Qualcomm Innovation Fellowship.

\bibliographystyle{unsrtnat} 
\bibliography{references}

\begin{thebibliography}{35}
\providecommand{\natexlab}[1]{#1}
\providecommand{\url}[1]{\texttt{#1}}
\expandafter\ifx\csname urlstyle\endcsname\relax
  \providecommand{\doi}[1]{doi: #1}\else
  \providecommand{\doi}{doi: \begingroup \urlstyle{rm}\Url}\fi

\bibitem[You et~al.(2018{\natexlab{a}})You, Liu, Ying, Pande, and
  Leskovec]{you2018graph}
Jiaxuan You, Bowen Liu, Rex Ying, Vijay Pande, and Jure Leskovec.
\newblock Graph convolutional policy network for goal-directed molecular graph
  generation.
\newblock In \emph{NeurIPS}, NIPS'18, page 6412–6422, Red Hook, NY, USA,
  2018{\natexlab{a}}. Curran Associates Inc.

\bibitem[Goyal et~al.(2020)Goyal, Jain, and Ranu]{goyal2020graphgen}
Nikhil Goyal, Harsh~Vardhan Jain, and Sayan Ranu.
\newblock Graphgen: a scalable approach to domain-agnostic labeled graph
  generation.
\newblock In \emph{Proceedings of The Web Conference 2020}, pages 1253--1263,
  2020.

\bibitem[De~Cao and Kipf(2018)]{molgan}
Nicola De~Cao and Thomas Kipf.
\newblock {MolGAN: An implicit generative model for small molecular graphs}.
\newblock \emph{ICML 2018 workshop on Theoretical Foundations and Applications
  of Deep Generative Models}, 2018.

\bibitem[Bojchevski et~al.(2018)Bojchevski, Shchur, Z{\"{u}}gner, and
  G{\"{u}}nnemann]{netgan}
Aleksandar Bojchevski, Oleksandr Shchur, Daniel Z{\"{u}}gner, and Stephan
  G{\"{u}}nnemann.
\newblock Netgan: Generating graphs via random walks.
\newblock In \emph{Proceedings of the 35th International Conference on Machine
  Learning, {ICML} 2018, Stockholmsm{\"{a}}ssan, Stockholm, Sweden, July 10-15,
  2018}, pages 609--618, 2018.

\bibitem[Albert and Barab{\'a}si(2002)]{albert2002statistical}
R{\'e}ka Albert and Albert-L{\'a}szl{\'o} Barab{\'a}si.
\newblock Statistical mechanics of complex networks.
\newblock \emph{Reviews of modern physics}, 74\penalty0 (1):\penalty0 47, 2002.

\bibitem[Liao et~al.(2019)Liao, Li, Song, Wang, Nash, Hamilton, Duvenaud,
  Urtasun, and Zemel]{gran}
Renjie Liao, Yujia Li, Yang Song, Shenlong Wang, Charlie Nash, William~L.
  Hamilton, David Duvenaud, Raquel Urtasun, and Richard Zemel.
\newblock Efficient graph generation with graph recurrent attention networks.
\newblock In \emph{NeurIPS}, 2019.

\bibitem[Simonovsky and Komodakis(2018)]{graphvae}
Martin Simonovsky and Nikos Komodakis.
\newblock Graphvae: Towards generation of small graphs using variational
  autoencoders.
\newblock In Vera Kurkov{\'{a}}, Yannis Manolopoulos, Barbara Hammer,
  Lazaros~S. Iliadis, and Ilias Maglogiannis, editors, \emph{Artificial Neural
  Networks and Machine Learning - {ICANN} 2018}, volume 11139 of \emph{Lecture
  Notes in Computer Science}, pages 412--422, 2018.

\bibitem[Gupta et~al.(2022)Gupta, Manchanda, Bedathur, and
  Ranu]{gupta2022tigger}
Shubham Gupta, Sahil Manchanda, Srikanta Bedathur, and Sayan Ranu.
\newblock Tigger: Scalable generative modelling for temporal interaction
  graphs.
\newblock In \emph{Proceedings of the AAAI Conference on Artificial
  Intelligence}, volume~36, pages 6819--6828, 2022.

\bibitem[Gupta and Bedathur(2022)]{gupta2022survey}
Shubham Gupta and Srikanta Bedathur.
\newblock A survey on temporal graph representation learning and generative
  modeling, 2022.

\bibitem[Jain et~al.(2021)Jain, Bagadia, Manchanda, and Ranu]{neuromlr}
Jayant Jain, Vrittika Bagadia, Sahil Manchanda, and Sayan Ranu.
\newblock Neuromlr: Robust \& reliable route recommendation on road networks.
\newblock \emph{Advances in Neural Information Processing Systems},
  34:\penalty0 22070--22082, 2021.

\bibitem[Borgwardt et~al.(2005)Borgwardt, Ong, Sch{\"o}nauer, Vishwanathan,
  Smola, and Kriegel]{bogwartenzymes}
Karsten~M. Borgwardt, Cheng~Soon Ong, Stefan Sch{\"o}nauer, S.~V.~N.
  Vishwanathan, Alexander~J. Smola, and Hans-Peter Kriegel.
\newblock Protein function prediction via graph kernels.
\newblock \emph{Bioinformatics}, 21 Suppl 1:\penalty0 i47--56, 2005.

\bibitem[National Center~for Biotechnology~Information()]{cancer}
U.S. National Library of~Medicine National Center~for
  Biotechnology~Information.
\newblock Pubchem.
\newblock URL \url{http://pubchem.ncbi.nlm.nih.gov}.

\bibitem[You et~al.(2018{\natexlab{b}})You, Ying, Ren, Hamilton, and
  Leskovec]{graphrnn}
Jiaxuan You, Rex Ying, Xiang Ren, William~L. Hamilton, and Jure Leskovec.
\newblock Graphrnn: Generating realistic graphs with deep auto-regressive
  models.
\newblock In \emph{ICML, 2018}, volume~80 of \emph{Proceedings of Machine
  Learning Research}, pages 5694--5703. {PMLR}, 2018{\natexlab{b}}.

\bibitem[Bartunov and Vetrov(2018)]{bartunov2018few}
Sergey Bartunov and Dmitry Vetrov.
\newblock Few-shot generative modelling with generative matching networks.
\newblock In \emph{International Conference on Artificial Intelligence and
  Statistics}, pages 670--678. PMLR, 2018.

\bibitem[Swinney and Xia(2014)]{swinney2014discovery}
David~C Swinney and Shuangluo Xia.
\newblock The discovery of medicines for rare diseases.
\newblock \emph{Future medicinal chemistry}, 6\penalty0 (9):\penalty0
  987--1002, 2014.

\bibitem[Cui et~al.(2020)Cui, Yang, and Yang]{10.3389/fmolb.2020.616341}
Wen Cui, Kailin Yang, and Haitao Yang.
\newblock Recent progress in the drug development targeting sars-cov-2 main
  protease as treatment for covid-19.
\newblock \emph{Frontiers in Molecular Biosciences}, 7, 2020.
\newblock ISSN 2296-889X.

\bibitem[Kipf et~al.(2018)Kipf, Fetaya, Wang, Welling, and
  Zemel]{kipf2018neural}
Thomas Kipf, Ethan Fetaya, Kuan-Chieh Wang, Max Welling, and Richard Zemel.
\newblock Neural relational inference for interacting systems.
\newblock In \emph{International Conference on Machine Learning}, pages
  2688--2697, 2018.

\bibitem[Perraudin et~al.(2019)Perraudin, Srivastava, Lucchi, Kacprzak,
  Hofmann, and R{\'e}fr{\'e}gier]{perraudin2019cosmological}
Nathana{\"e}l Perraudin, Ankit Srivastava, Aurelien Lucchi, Tomasz Kacprzak,
  Thomas Hofmann, and Alexandre R{\'e}fr{\'e}gier.
\newblock Cosmological n-body simulations: a challenge for scalable generative
  models.
\newblock \emph{Computational Astrophysics and Cosmology}, 6\penalty0
  (1):\penalty0 1--17, 2019.

\bibitem[Yuanqi(2021)]{du2021graphgt}
Yuanqi.
\newblock Graphgt: Machine learning datasets for graph generation and
  transformation.
\newblock In \emph{Thirty-fifth Conference on Neural Information Processing
  Systems Datasets and Benchmarks Track (Round 2)}, 2021.

\bibitem[Irwin and Shoichet(2005)]{irwin2005zinc}
John~J Irwin and Brian~K Shoichet.
\newblock Zinc- a free database of commercially available compounds for virtual
  screening.
\newblock \emph{Journal of chemical information and modeling}, 45\penalty0
  (1):\penalty0 177--182, 2005.

\bibitem[Finn et~al.(2017)Finn, Abbeel, and Levine]{finn2017model}
Chelsea Finn, Pieter Abbeel, and Sergey Levine.
\newblock Model-agnostic meta-learning for fast adaptation of deep networks.
\newblock In \emph{International conference on machine learning}, pages
  1126--1135. PMLR, 2017.

\bibitem[Manchanda et~al.(2022)Manchanda, Michel, Drakulic, and
  Andreoli]{manchanda2022generalization}
Sahil Manchanda, Sofia Michel, Darko Drakulic, and Jean-Marc Andreoli.
\newblock On the generalization of neural combinatorial optimization
  heuristics.
\newblock In \emph{Joint European Conference on Machine Learning and Knowledge
  Discovery in Databases}, pages 426--442. Springer, 2022.

\bibitem[Yan(2002)]{gspan}
Xifeng Yan.
\newblock gspan: Graph-based substructure pattern mining.
\newblock In \emph{Proceedings of the 2002 {IEEE} International Conference on
  Data Mining {(ICDM} 2002), 9-12 December 2002, Maebashi City, Japan}, pages
  721--724, 2002.

\bibitem[Hochreiter and Schmidhuber(1997)]{10.1162/neco.1997.9.8.1735}
Sepp Hochreiter and Jürgen Schmidhuber.
\newblock {Long Short-Term Memory}.
\newblock \emph{Neural Computation}, 9\penalty0 (8):\penalty0 1735--1780, 11
  1997.
\newblock ISSN 0899-7667.

\bibitem[Nichol and Schulman(2018)]{nichol2018reptile}
Alex Nichol and John Schulman.
\newblock Reptile: a scalable metalearning algorithm.
\newblock \emph{arXiv preprint arXiv:1803.02999}, 2\penalty0 (3):\penalty0 4,
  2018.

\bibitem[Zaremba and Sutskever(2015)]{zaremba2015learning}
Wojciech Zaremba and Ilya Sutskever.
\newblock Learning to execute, 2015.

\bibitem[Kumar et~al.(2010)Kumar, Packer, and Koller]{NIPS2010_e57c6b95}
M.~Kumar, Benjamin Packer, and Daphne Koller.
\newblock Self-paced learning for latent variable models.
\newblock In \emph{NeurIPS}, volume~23. Curran Associates, Inc., 2010.

\bibitem[Kawai et~al.(2019)Kawai, Mukuta, and Harada]{gram}
Wataru Kawai, Yusuke Mukuta, and Tatsuya Harada.
\newblock {GRAM:} scalable generative models for graphs with graph attention
  mechanism.
\newblock \emph{CoRR}, abs/1906.01861, 2019.

\bibitem[Ho{\v{c}}evar and Dem{\v{s}}ar(2014)]{hovcevar2014combinatorial}
Toma{\v{z}} Ho{\v{c}}evar and Janez Dem{\v{s}}ar.
\newblock A combinatorial approach to graphlet counting.
\newblock \emph{Bioinformatics}, 30\penalty0 (4):\penalty0 559--565, 2014.

\bibitem[Gretton et~al.(2012)Gretton, Borgwardt, Rasch, Sch{\"{o}}lkopf, and
  Smola]{metric:mmd}
Arthur Gretton, Karsten~M. Borgwardt, Malte~J. Rasch, Bernhard Sch{\"{o}}lkopf,
  and Alexander~J. Smola.
\newblock A kernel two-sample test.
\newblock \emph{J. Mach. Learn. Res.}, 13:\penalty0 723--773, 2012.

\bibitem[Costa(2010)]{nspdk}
Fabrizio Costa.
\newblock Fast neighborhood subgraph pairwise distance kernel.
\newblock In \emph{ICML}, pages 255--262, 2010.

\bibitem[Ingraham et~al.(2019)Ingraham, Garg, Barzilay, and
  Jaakkola]{ingraham2019generative}
John Ingraham, Vikas Garg, Regina Barzilay, and Tommi Jaakkola.
\newblock Generative models for graph-based protein design.
\newblock \emph{Advances in Neural Information Processing Systems}, 32, 2019.

\bibitem[Guo et~al.(2020)Guo, Du, Tadepalli, Zhao, and
  Shehu]{guo2020generating}
Xiaojie Guo, Yuanqi Du, Sivani Tadepalli, Liang Zhao, and Amarda Shehu.
\newblock Generating tertiary protein structures via an interpretative
  variational autoencoder.
\newblock \emph{arXiv preprint arXiv:2004.07119}, 2020.

\bibitem[Schomburg et~al.(2004)Schomburg, Chang, Ebeling, Gremse, Heldt, Huhn,
  and Schomburg]{brendaenzymes}
Ida Schomburg, Antje Chang, Christian Ebeling, Marion Gremse, Christian Heldt,
  Gregor Huhn, and Dietmar Schomburg.
\newblock Brenda, the enzyme database: updates and major new developments.
\newblock \emph{Nucleic acids research}, 32 Database issue:\penalty0 D431--3,
  2004.

\bibitem[Vaswani et~al.(2017)Vaswani, Shazeer, Parmar, Uszkoreit, Jones, Gomez,
  Kaiser, and Polosukhin]{NIPS2017_3f5ee243}
Ashish Vaswani, Noam Shazeer, Niki Parmar, Jakob Uszkoreit, Llion Jones,
  Aidan~N Gomez, \L~ukasz Kaiser, and Illia Polosukhin.
\newblock Attention is all you need.
\newblock In I.~Guyon, U.~Von Luxburg, S.~Bengio, H.~Wallach, R.~Fergus,
  S.~Vishwanathan, and R.~Garnett, editors, \emph{Advances in Neural
  Information Processing Systems}, volume~30. Curran Associates, Inc., 2017.
\newblock URL
  \url{https://proceedings.neurips.cc/paper_files/paper/2017/file/3f5ee243547dee91fbd053c1c4a845aa-Paper.pdf}.

\end{thebibliography}

\clearpage

\section{Appendix}
\appendix

\section{Notations}

\begin{table}[h]
\label{tab:notations}
    \centering
    \scalebox{1}{
        \begin{tabular}{m{0.3\linewidth}  m{0.6 \linewidth}}\\
         \toprule
        \textbf{Symbol} & \textbf{Meaning}\\\midrule
        $G = (V, E)$ &  A Graph with vertex set $V$ and edge set $E$ \\    \cmidrule(lr){1-2}
         $n$& Number of nodes in $G$\\
         \cmidrule(lr){1-2}
         $m$& Number of edges in $G$\\
         \cmidrule(lr){1-2}
         $\mathbb{V}$& Label set of vertices in $G$\\
         \cmidrule(lr){1-2}
         $\mathbb{E}$ & Label set of edges in $G$ \\
         \cmidrule(lr){1-2}
        $D{=}\{G_1, G_2,\cdots, G_N\}$ & Dataset of $N$ graphs\\
        \cmidrule(lr){1-2}
        $\mathcal{D}{=}\{D_1,\cdots, D_{\mathcal{B}} \}$ & Collection of $\mathcal{B}$ graph datasets \\
        \cmidrule(lr){1-2}
        $t_u$ & DFS discovery time of node $u$\\
        \cmidrule(lr){1-2}
        $t_v$ & DFS discovery time of node $v$ \\
        \cmidrule(lr){1-2}
        $L_{u}$ & Label of node $u$ \\
        \cmidrule(lr){1-2}
        $L_{uv}$ & Label of edge $(u,v)$ \\
        \cmidrule(lr){1-2}
        $L_v$ & Label of node $v$ \\

        \cmidrule(lr){1-2}
        $\mathcal{F(G)}$ & Function to map graph to Minimum DFS code $S$\\
                \cmidrule(lr){1-2}
        $S=(s_1,s_2\ldots s_m)$ & Minimum DFS Codes of a graph \\ \cmidrule(lr){1-2} $\mathcal{S} =\{S_1, S_2\cdots  S_N\}$  & Collection of Minimum DFS codes of a dataset with $N$ graphs  \\
        \cmidrule(lr){1-2}
         
        $\mathcal{T}$ & Set of graph generative modelling tasks \\
        \cmidrule(lr){1-2}
        $\mathcal{T}_i$ & Graph generative modelling task for the $i^{th}$ dataset \\
        
        \cmidrule(lr){1-2}
         $\mathcal{L}_{i}$  & Loss associated with  dataset $D_i$\\
        \cmidrule(lr){1-2}
        $\theta$ & Model parameters \\
        \cmidrule(lr){1-2}
       
        $D_T$ & Target graph dataset \\
        \cmidrule(lr){1-2}
         \cmidrule(lr){1-2}
         $\mathcal{S}_{T}$  & Collection of Minimum DFS codes for target dataset $D_T$\\
        
        \cmidrule(lr){1-2}
        $\theta_{T}$  & Parameters fine-tuned to the  dataset $D_T$ \\
        \cmidrule(lr){1-2}
        $\beta_i$  & Binary loss coefficient for $i^{th}$ sample in eq. \ref{eq:self_pace_loss} \\
        \cmidrule(lr){1-2}
        $\gamma$ & Growth parameter in self-paced fine-tuning \\ 
        \bottomrule
        \end{tabular}
        }
    \caption{Notations used in the paper}
    \label{tab:notation}
\end{table}


\newpage
\section{Pseudocodes}

\begin{algorithm}[h!]
\DontPrintSemicolon
\scriptsize
\SetKwInOut{Input}{Input}
\SetKwInOut{Output}{Output}
\Input{Collection of $\mathcal{B}$ graph datasets $\mathcal{D}=\{D_1,D_2 \ldots D_{\mathcal{B}}\}$, $K$, $\epsilon$}
\Output{Good initialization of parameters $\theta$ of generative model $p_{\theta}(D)$}
Initialise meta-parameters $\theta$ randomly.\;

\Repeat(){stopping criteria}{
 Sample a dataset $D \in \mathcal{D}$\;
 
 $\mathcal{S}=\{S= \mathcal{F}(G)\:|\: \forall G\in D\}$ \tcp*{Get Minimum DFS code}
 $\theta_D \leftarrow \theta$
\tcp*{Dataset $D$ specific parameters }
\For(\tcp*[f]{$K$ inner gradient steps }) {$K$ times}{
      $S = [s_1,s_2 \ldots s_m] \sim \mathcal{S}$\;
      $s_0$ $\leftarrow$ SOS\;
      
      $h_0$ $\leftarrow$ $\boldsymbol{0}$ \;
      $\mathcal{L}_{D} \leftarrow 0$\;
      \tcc{Computing loss $\mathcal{L}_D$ of sequence $S=[s_0,s_1,s_2 \ldots s_{m+1}]$}
        \For(\tcp*[f]{$s_{m+1}$ for EOS tokens}){ $i$ from $1$ to $m + 1$}{
          $\ch_i \leftarrow LSTM^{hidden}_{\theta}(\ch_{i-1}, f^{emb}_{\theta}(s_{i-1}))$\;
        
         $\widetilde{s}_i \leftarrow \left\langle f^{t_u}_{\theta}(\ch_i),f^{t_v}_{\theta}(\ch_i),f^{L_u}_{\theta}(\ch_i),f^{L_{uv}}_{\theta}(\ch_i),f^{L_v}_{\theta}(\ch_i)\right\rangle$ \;
         
         $\mathcal{L}_{D} \leftarrow \mathcal{L}_{D} + \sum_{c}\left(s_{i}[c] \log \widetilde{s}_{i}[c]+\left(1-s_{i}[c]\right) \log \left(1-\widetilde{s}_{i}[c]\right)\right)$\;
          
        }
         $\theta_{D} \leftarrow \theta_{D}-\alpha \nabla_{\theta_{D}} \mathcal{L}_{D}$ \tcp*{D specific parameters' update}
        
  }
  Update $\theta \leftarrow \theta + \epsilon(\theta_D - \theta)$ \tcp*{Meta gradient update}
}(\tcp*[f]{Typically when validation loss is minimized})
\caption{Pseudocode for meta-training phase of  \namemodel}
\label{alg:meta_training}
\end{algorithm}

\begin{algorithm}[h!]
\DontPrintSemicolon
\scriptsize
\SetKwInOut{Input}{Input}
\SetKwInOut{Output}{Output}
\Input{Target dataset $D_T$, meta-trained parameters $\theta$, batch size $B$, growth factor $\gamma$, $\lambda$}
\Output{Fine tuned parameters $\theta_{T}$ for target dataset $D_T$}
 $\mathcal{S_T}=\{S= \mathcal{F}(G)\:|\: \forall G\in D_T\}$\tcp*{Get Minimum DFS code}
 $\theta_{T} \leftarrow \theta$ \tcp*{Initializing parameters specific to target dataset $D_T$}
 
\Repeat(){stopping criteria}{
$\mathcal{L}_{T} \leftarrow 0$;

\For(\tcp*[f]{Sample $B$ graphs for every batch}){$B$ times}{
       $S = [s_1,s_2 \ldots s_m] \sim \mathcal{S_T}$\;
       $s_0$ $\leftarrow$ SOS\;
      
       $h_0$ $\leftarrow$ $\boldsymbol{0}$ \;
       
       $l \leftarrow 0$ \tcp*{Instance specific loss}
             \tcc{Computing loss $l$ of sequence $S=[s_0,s_1,s_2 \ldots s_{m+1}]$}
        \For(\tcp*[f]{$s_{m+1}$ for EOS tokens}){ $i$ from $1$ to $m + 1$}{
           $\ch_i \leftarrow LSTM^{hidden}_{\theta_{T}}(\ch_{i-1}, f^{emb}_{\theta_{T}}(s_{i-1}))$\;
        
         $\widetilde{s}_i \leftarrow (f^{t_u}_{\theta_{T}}(\ch_i),f^{t_v}_{\theta_{T}}(\ch_i),f^{L_u}_{\theta_{T}}(\ch_i),f^{L_{uv}}_{\theta_{T}}(\ch_i),f^{L_v}_{\theta_{T}}(\ch_i))$ \;
         
         $l \leftarrow l + \sum_{c}\left(s_{i}[c] \log \widetilde{s}_{i}[c]+\left(1-s_{i}[c]\right) \log \left(1-\widetilde{s}_{i}[c]\right)\right)$\;
          
        }
        \uIf{$l < \lambda $}{
            $\mathcal{L}_{T} = \mathcal{L}_{T} + l$ \;
        }
        
  }
         $\theta_{T} \leftarrow \theta_{T}-\alpha \nabla_{\theta_{T}}\mathcal{L}_{T}$ \;
         
         $\lambda = \lambda*\gamma$ \tcp*{Increase difficulty  periodically}
}(\tcp*[f]{Typically when validation loss is minimized})
\caption{Fine-tuning \namemodel on target dataset $D_T$}
\label{alg:fine_tuning}
\end{algorithm}


\begin{algorithm}[h!]
\DontPrintSemicolon
\scriptsize
\SetKwInOut{Input}{Input}
\SetKwInOut{Output}{Output}
\Input{Fine tuned parameters $\theta_{T}$ of deep generative model $p_{\theta_{T}}$}
\Output{Graph $G$}
$S \leftarrow \left(\right)$\; 
$s_0 \leftarrow SOS$\; \label{inf:init:s}
$i \leftarrow 0$\; 
$\ch_0 \leftarrow \boldsymbol{0}$\;
\Repeat{EOS $\in \{s_i.t_u, s_i.t_v, s_i.L_u, s_i.L_{uv}, s_i.L_v$\} }{
$i \leftarrow i + 1$ \;
$\ch_i \leftarrow LSTM^{hidden}_{\theta_{T}}(\ch_{i-1}, f^{emb}_{\theta_{T}}(s_{i-1}))$\;
\tcp{Sample $s_i$ from multinomial distributions parameterized using $\ch_i$}
$s_i.t_u \sim Multinomial(f^{t_u}_{\theta_{T}}(\ch_i))$\;
$s_i.t_v \sim  Multinomial(f^{t_v}_{\theta_{T}}(\ch_i))$ \;
$s_i.L_u \sim Multinomial(f^{L_u}_{\theta_{T}}(\ch_i)) $\;
$s_i.L_{uv} \sim Multinomial(f^{L_{uv}}_{\theta_{T}}(\ch_i)) $\;
$s_i.L_v \sim  Multinomial(f^{L_v}_{\theta_{T}}(\ch_i)) $\;
$S.append(s_i)$\;
}(\tcp*[f]{Check if any item of tuple $s_i$ contains $EOS$ symbol})
$G \leftarrow {\mathcal{F}}^{-1}(S)$ \tcp*{Convert DFScode back to graph}

\Return{$G$}
\caption{Pseudocode of graph generation for target dataset $D_T$}
\label{alg:graph_generation}
\end{algorithm}

\newpage

\section{Dataset Semantics}
\label{app:dataset-desc}
    \noindent \textbf{Biological Domain:} Proteins are biomolecules consisting of long chain of amino acids. They are highly essential to our lives and significantly interesting in certain biomedical tasks such as \textit{de novo} protein design\cite{ingraham2019generative,guo2020generating}. 
     Enzymes, a set of  specialized proteins, are catalysts that can speed up metabolic activities. In our work, we utilize the Enzyme dataset from the BRENDA enzyme database~\cite{brendaenzymes}, which consists of protein tertiary structures.
    We convert enzymes to graphs where nodes represent secondary structures labeled into one of the three categories namely \textit{helices}, \textit{turns}, or \textit{sheets}. This dataset does not have edge labels. The dataset is divided into six classes and each enzyme belongs to one of these classes, namely EC1, EC2, EC3, EC4, EC5, EC6. For our limited data learning setup, we consider learning to generate graphs belonging to a certain enzyme class as a task. We treat the datasets EC1, EC2, EC4, EC5, EC6 as auxiliary and EC3 as our target dataset, which consists of $100$ enzymes. 
    
    \noindent \textbf{Chemical Domain:} Chemical compounds  are composed of two or more atoms connected using chemical bonds.  We utilize the following chemical compounds datasets to train and evaluate \namemodel.
    
    \noindent \textit{AIDS-CA~\cite{gspan}:} This dataset comprises of a set of molecules that displayed activity against HIV. 

    \noindent \textit{Breast, Lung, Yeast:} Each of these three datasets contain molecules that were screened for activity against Breast cancer, Lung cancer and cancer in Yeast respectively~\cite{cancer}.

    
    
    \noindent \textit{Leukemia-Active:} This dataset consists of  compounds that are active against Leukemia~\cite{cancer}.
    
    
    In all chemical datasets, we convert compounds to labeled graphs where nodes represent atoms and their labels represent atom-type which are elements belonging to the chemical periodic table. Edges in the graphs represent bonds and edge labels encode the bond type i.e single, double, triple.
    
    For the limited data learning setup for chemical domain, \textit{Yeast}, \textit{Breast}, and \textit{Lung} are used as auxiliary datasets during meta-training. Further, we choose AIDS-CA and Leukemia-Active datasets as our target datasets. The reasons for this choice is \textbf{(1)} due to their relatively low availability of number of graph samples and \textbf{(2)} since they consist of compounds that are active against certain diseases, therefore have more practical utility.
    

    \noindent \textbf{Physics Domain:} 
    Physics-based simulations are commonly used to understand interactions among different objects\cite{du2021graphgt,perraudin2019cosmological, kipf2018neural}. Dynamical systems such as $N$-body springs can be converted into graph structures where nodes represent particles and edges represent connections between particles. We utilize the dataset of the $N$-body spring simulations~\cite{du2021graphgt}. It consists of $N$ particles in a two-dimensional space partitioned into a $5{\times}5$ grid . Two particles are connected to each other via spring with a probability of $0.5$. The label of a node is the partition it lies in. 
     This system does not have edge labels\cite{du2021graphgt}. For  learning under limited data, we meta-train \namemodel on auxiliary datasets consisting of four and six particle systems and then fine-tune on graphs containing five particles.
\begin{table}[t]
 \vspace{-0.30in}
 \centering
\caption{Summary of the datasets} 
\label{tab:datasets}
{\scriptsize
  \begin{tabular}{lllccccc}
    \toprule
    \# & Name & Domain & No. of graphs & $|V|$ & $|E|$ & $|\mathbb{V}|$ & $|\mathbb{E}|$\\
    \midrule
    1&Enzymes\cite{bogwartenzymes}& Biological & 600 & [2, 125] & [2, 149] & 3 & X  \\
    2& NCI-H23 (Lung)\cite{cancer}& Chemical & 24k & [6, 50] & [6, 57] & 11 & 3 \\
    5& Yeast\cite{cancer}& Chemical & 47k & [5, 50] & [5, 57] & 11 & 3 \\
    7&MCF-7 (Breast)\cite{cancer}& Chemical & 23k & [6, 111] & [6, 116] & 11 & 3 \\
    6& Leukemia-Active\cite{cancer} & Chemical & 1900 & [12, 107] & [12, 111] & 11 & 3 \\
    6& AIDS-CA\cite{cancer} & Chemical & 328 & [10, 189] & [10, 196] & 11 & 3 \\

     8&N-body Spring\cite{du2021graphgt}& Physics & 1500 & N & [3, 13] & 25 & X \\
  \bottomrule
\end{tabular}}
 \vspace{-0.2in}
\end{table}


\section{Experimental Setup and Reproducibility}
\label{sec:app:exp}
All experiments are performed on a machine with Intel Xeon Gold 6284 processor with 96 physical cores, 1 NVIDIA A100 GPU card with 40GB GPU memory, and 512 GB RAM running Ubuntu 20.04 operating system.   

 
\subsection{Parameter details}
\label{app:parameters}
We set hidden dimension of $f^{L_u}_{\theta}$,
$f^{L_v}_{\theta}$, $f^{L_{uv}}_{\theta}$, $f^{t_u}_{\theta}$, $f^{t_v}_{\theta}$ to $512$. We utilize \textit{Adam optimizer} with learning rate as $0.003$. Further to avoid over-fitting we use dropout with value of $0.2$ and an L2 regularizer with value of $10^{-5}$. We set batch size to $32$. For meta-training of \namemodel we used $K{=}15$ and $\epsilon=0.8$. For Enzyme we set $K=50$ and $\epsilon=0.5$.  During fine-tuning, we used the value of the growth-factor $\gamma{=}1.001$ for both Leukemia-Active and Enzyme,  $1.006$ for AIDS-CA and $1.1$ for 5-body spring. For all methods, we stop training when validation loss is minimized or there is less than $0.05\%$ change in validation loss over a number of extended epochs.

\subsection{Number of graphs generated}
\label{sec:app:graphs_gen}
Since our test datasets are of different sizes, AIDS-CA ($108$), Leukemia-Active ($900$), Enzyme-EC3 ($20$), 5-body Spring ($500$), we generate a different number of graphs for each target dataset. Specifically, for Leukemia-Active we generate $2560$ graphs, $1024$ graphs each for AIDS-CA and $5$-body spring, and for Enzyme EC3 we generate $512$ graphs.

\section{Impact of Self-paced fine tuning}
\label{app:self_pace}
\begin{figure}[h!]
\centering
\includegraphics[scale=0.18]{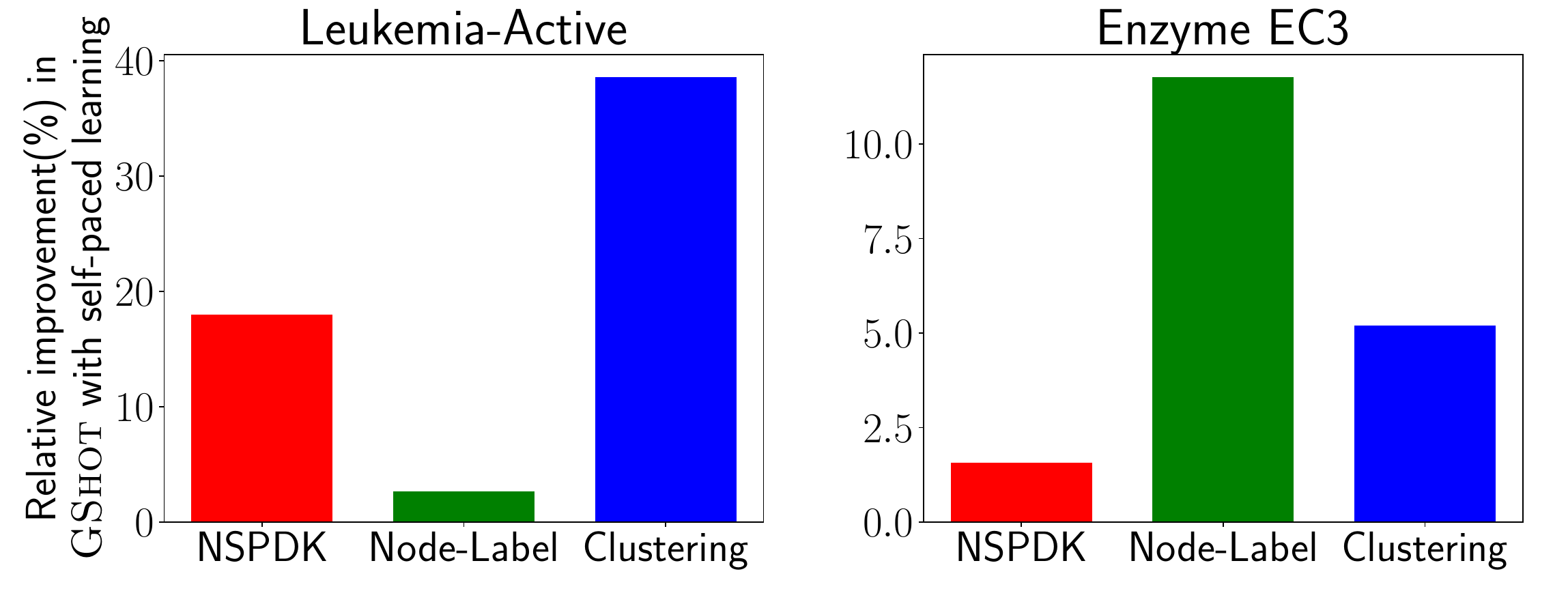}  
  \vspace{-0.05in}
 \caption{ \label{fig:ablation} Ablation study showing the relative (\%) improvement obtained by \namemodel when using \textit{self-paced fine-tuning} compared to \namemodel with \textit{vanilla fine-tuning.}}
\end{figure}

We study the improvement obtained by using \textit{self-paced fine-tuning} in  \namemodel over {\textit{vanilla fine-tuning}} on different metrics. For a metric $P$, we define the improvement as   $\frac{P_{\namemodel(vanilla)} -  P_{\namemodel} }{P_\namemodel}\times 100$. Here, $P_{\namemodel}$ refers to the value of the metric $P$ obtained by our default model (with self-paced fine tuning), and $P_{\namemodel(vanilla)}$ refers to the value obtained by \namemodel with vanilla fine-tuning. In Fig.~\ref{fig:ablation} we observe that a self-paced fine-tuning strategy can improve the fidelity metrics significantly.

\section{Impact of auxiliary datasets}
In this section, we study the performance of our proposed architecture by selecting different auxiliary datasets during meta-training. Towards this, we choose Enzyme dataset since it consists of $5$ auxiliary datasets and has reasonable scope to sample \textit{multiple sets of auxiliary datasets} from it. For this experiment we sample(without repetition) sets of $3$ auxiliary datasets $5$ times(eg:- \{EC1, EC4, EC5\}, \{EC2, EC5, EC6\} etc.). We train $5$ \namemodel models  with these $5$  different sets of auxiliary datasets. We then fine-tune these $5$ trained models on the target dataset(EC3). We use the same set of auxiliary datasets for training the \textsc{PreTrain+FT} baseline.
In Table~\ref{tab:quality:aux}, we report the mean performance on each metric along with standard deviation obtained using these $5$ models. For results of training \graphGen and \graphRNN from scratch directly on the target dataset EC3(without auxiliary datasets), refer to  Table~\ref{tab:quality} in the main paper.

\label{var:aux:dataset}

\renewcommand{\tabcolsep}{3pt}
\begin{table*}[h]
\caption{\textbf{Performance on variation of auxiliary datasets:} Performance comparison on the Enzyme EC3 dataset when different sets of auxiliary datasets are used for meta-training \namemodel and for training the  \textsc{PreTrain+FT} baseline. For \namemodel and \textsc{PreTrain+FT}, we report the mean and standard deviation since their performance is averaged across models using different sets of auxiliary datasets used for (meta/pre) training.}
\label{tab:quality:aux}\hspace{-0.1in}
\hspace{-0.1in}
\scalebox{0.85}{
{\scriptsize
\begin{tabular}{c|c|ccc|c|cc|ccc|cc} 
\toprule 
   \textbf{Target dataset}&\textbf{Model} & \textbf{Deg.} & \textbf{Clus.} & \textbf{Orbit} & \textbf{NSPDK}  & \makecell{\\\textbf{Avg} \# \textbf{Nodes} \\ \textbf{(Gen/Gold)}} & \makecell{\\\textbf{Avg} \# \textbf{Edges} \\ \textbf{(Gen/Gold)}} & \makecell{\\\textbf{Node} \\ \textbf{Label}} & \makecell{\\\textbf{Edge} \\ \textbf{Label}} & \makecell{\\\textbf{Joint}  \textbf{Node}  \\ \textbf{Label}  \& \textbf{Degree}} & \textbf{Novelty} & \textbf{Uniqueness}  \\
    
    \midrule
    
     \multirow{2}{5em}{Enzyme:EC3 \\ \#Training samples=50} & {\makecell{\textsc{PreTrain+FT}  }} & 0.731 &	0.585	& 0.060 &	0.191 &	20.596/26.90 &	28.658/52.85 &	0.006	& {x}	 & 0.644	 & 98.50\%	 &98.70\% \\

     & &$\pm$0.077&$\pm$0.066&$\pm$0.012&$\pm$0.007&$\pm$1.150&$\pm$1.265&$\pm$0.002&x&$\pm$0.026&$\pm$0.005&$\pm$0.004
    \\ 
\cline{2-13}
& \multirow{2}{*}{\makecell{\namemodel  }} & 0.710 &	0.554 & 	0.053 &	0.188 &	20.742/26.90 &	29.212/52.85 &	0.004 & x	&	0.634 &	 98.11\% &	98.12\%
\\

 &  & $\pm$0.07 &	$\pm$0.067 &	$\pm$0.009 &$\pm$	0.005 &$\pm$	1.04 &$\pm$	1.27	&$\pm$ 0.001 & x &		$\pm$0.018 & $\pm$	0.008&$\pm$	0.009
\\
 
    \bottomrule
\end{tabular}}
}
\end{table*}

In Table~\ref{tab:quality:aux}, we observe that \namemodel obtains superior performance when trained using different sets of auxiliary datasets. 
For instance, on the \textit{Node label} metric, \namemodel outperforms its closest competitor \textsc{PreTrain+FT} by around $50\%$. Further, it outperforms its closest competitor by over $10\%$ on the \textit{Orbit} metric. 
Overall, we observe that \namemodel learns to better utilize the knowledge gained from a variety of auxiliary datasets.

\section{Learning to conditionally predict elements of  edge tuple of DFS code }
\label{app:cond_edge_tuple}
 We assumed independence between components of edge tuple in order to simplify the modeling process. We also performed experiments by removing the independence assumption and making the prediction of components of edge tuple $\left\langle t_u, t_v,L_u, L_{uv},L_v\right\rangle$ conditional on other elements. In order to model the conditional distribution, an order has to be imposed on the components of an edge tuple $\left\langle t_u, t_v,L_u, L_{uv},L_v\right\rangle$. As discussed in Sec. 3.2, in the generation process of DFS code for a graph, first a node is discovered and consequently its discovery time $t_u$ is recorded. Since this component depends upon the hidden state of the DFS code sequence, we first predict the timestamp $t_u$ of the edge tuple using the hidden state $h_i$ of LSTM. Now, in order to form an edge, the node with timestamp $t_u$ is to be connected to node with discovery timestamp $t_v$. Since $t_v$ depends upon the hidden state $h_i$  and the timestamp $t_u$ of the discovered node, we concatenate the hidden state embedding $h_i$ and embedding $emb(t_u)$ of the timestamp $t_u$ and predict the timestamp $t_v$. Now, for the remaining three components $L_u$ , $L_v$ and $L_{uv}$, we first predict the label of $u$ and then $v$ in order. This choice promises to improve likelihood of generating node labels $L_v$ that have high probability of forming edges with nodes having label $L_u$ and vice-versa. Finally, since the edge label between two nodes is a function of the node type(labels), therefore, we condition  the edge label $L_{uv}$ on the embedding of labels $L_u$ and $L_v$ of the edge tuple. The details of the conditional modeling are presented below:
\begin{align}
\nonumber
   \label{eq:cond}
  \nonumber  t_u &=MLP(\ch_i) \\
  \nonumber  t_v &=MLP(\ch_i|emb(t_u))\\
  \nonumber  L_u &=MLP(\ch_i|emb(t_u)|emb(t_v)) \\
  \nonumber  L_v &=MLP(\ch_i|emb(t_u)|emb(t_v)|emb(L_u)) \\
  \nonumber  L_{uv}&=MLP(\ch_i|emb(t_u)|emb(t_v)|emb(L_u)|emb(L_v))
\end{align}
     
Here $emb$ refers to learnable embedding layer whose parameters are learnt during the training process. 

\begin{table}[h!]
\centering
\scalebox{0.85}{
{\scriptsize
\begin{tabular}{p{7em}|p{7em}|p{3.8em}|c|ccc|c|ccc|cc} 

\textbf{Target dataset} & \textbf{Model} & \textbf{Deg.} & \textbf{Clus.} & \textbf{Orbit} & \textbf{NSPDK}  & \makecell{\textbf{Avg} \# \textbf{Nodes} \\ \textbf{(Gen/Gold)}} & \makecell{\textbf{Avg} \# \textbf{Edges} \\ \textbf{(Gen/Gold)}} & \makecell{\textbf{Node} \\ \textbf{Label}} & \makecell{\textbf{Edge} \\ \textbf{Label}} & \makecell{\textbf{Joint} \\ \textbf{Node} \\ \textbf{Label} \\ \& \textbf{Degree}} & \textbf{Novelty} & \textbf{Uniqueness}   \\ \hline

\makecell{ Leukemia-Active} & \namemodel(Ind) & ${0.0069}$ & $\approx$ 0 & $\approx$ 0 & 0.032 & 42.35/47.71 & 44.33/50.37 & 0.0011 & $\approx$ 0 & 0.24 & 100\% & 100\% \\ 
 & \namemodel(Cond) & ${0.01}$ & $\approx$ 0.003 & $\approx$ 0 & 0.06 & 46.37/47.71 & 49.23/50.37 & 0.003 & $\approx$ 0 & 0.44 & 100\% & 100\% \\ \cline{1-13}
 \makecell{ AIDS-CA} & \namemodel(Ind) & 0.017 & 0.0015 & $\approx$ 0 & 0.08 & 26.5/37.14 & 27.1/39.60 & 0.011 & $\approx$ 0 & 0.14 & 99\% & 99\% \\
 & \namemodel(Cond) & 0.01 & 0.004 & $\approx$ 0 & 0.07 & 28.44/37.14 & 30.50/39.60 & 0.007 & $\approx$ 0 & 0.12 & 98.33\% & 98.43\% \\ \cline{1-13}
 \makecell{ Enzyme EC3} & \namemodel(Ind) & 0.45 & 0.47 & 0.025 & 0.16 & 24.5/26.90 & 37.69/52.85 & 0.004 & x & 0.457 & 100\% & 100\% \\ 
 & \namemodel(Cond) & 0.43 & 0.60 & 0.03 & 0.20 & 27.50/26.90 & 42.55/52.85 & 0.007 & x & 0.58 & 99.4\% & 98.8\% \\ \cline{1-13}
\end{tabular}

}}
\caption{\textbf{Conditional vs Independent modeling of edge tuple: } Performance obtained by different models on different datasets on multiple metrics when components of edge tuples are predicted independently(Ind) vs when they are predicted conditionally(Cond).\label{tab:cond_ind_edge}}
\end{table}
     
To evaluate the performance of the above conditional prediction of the edge tuple, we ran experiments on the target dataset Leukemia-Active, Enzyme EC3 and AIDS-CA dataset . The results are presented in table~\ref{tab:cond_ind_edge}. We denote the model with Independence assumption(default) as Ind, and the model with conditional prediction of components in edge tuple as Cond. On all datasets we observe that for the metrics Avg \#nodes and Avg \#edges, the models with conditional modeling of edge tuple convincingly outperforms the model(s) with the independence assumption. This signifies that sequence length has been better modelled in \namemodel when components of  the edge tuple are modeled conditionally in comparison to when modeled independently. However, on the other distance metrics, we do not observe clear performance improvements.

\section{Using transformer for sequence Modelling}
We integrated transformer \cite{NIPS2017_3f5ee243} for modeling the sequences. We present the results in the table ~\ref{tab:trans_lstm} when compared with current LSTM-based architecture.

On all datasets, we do not observe clear performance improvements.

\begin{table}[h!]
\centering
\scalebox{0.95}{
{\scriptsize
\resizebox{\textwidth}{!}{
\begin{tabular}{p{7.5em}|p{10em}|p{3.8em}|c|ccc|c|ccc|cc}

\textbf{Target dataset} & \textbf{Model} & \textbf{Deg.} & \textbf{Clus.} & \textbf{Orbit} & \textbf{NSPDK}  & \makecell{\textbf{Avg} \# \textbf{Nodes} \\ \textbf{(Gen/Gold)}} & \makecell{\textbf{Avg} \# \textbf{Edges} \\ \textbf{(Gen/Gold)}} & \makecell{\textbf{Node} \\ \textbf{Label}} & \makecell{\textbf{Edge} \\ \textbf{Label}} & \makecell{\textbf{Joint} \\ \textbf{Node} \\ \textbf{Label} \\ \& \textbf{Degree}} & \textbf{Novelty} & \textbf{Uniqueness}  \\ \hline
Leukemia- Active & {\namemodel} (LSTM) & ${0.0069}$ & $\approx$ 0 & $\approx$ 0 & 0.032 & 42.35/47.71 & 44.33/50.37 & 0.0011 & $\approx$ 0 & 0.24 & 100 & 100 \\ 
& {\namemodel}(Transformer) & $0.0256$ & $\approx$ 0.0067 & $\approx$ 0.0018 & 0.1983 & 50.22/47.71 & 54.57/50.37 & 0.011 & $\approx$ 0.00104 & 0.522 & 100 & 99.85 \\ 
\hline
AIDS-CA & \namemodel (LSTM) & 0.017 & 0.0015 & $\approx$ 0 & 0.08 & 26.5/37.14 & 27.1/39.60 & 0.011 & $\approx$ 0 & 0.14 & 99\% & 99\% \\ 
&{\namemodel} (Transformer) & 0.03044 & 0.001543 & 0.000680 & 0.100 & 27.10/37.14 & 28.64/39.60 & 0.02865 & 0.00089 & 0.2220 & 100\% & 99.70\% \\ 
\hline
Enzyme EC3  & \namemodel (LSTM) & 0.45 & 0.47 & 0.025 & 0.16 & 24.5/26.90 & 37.69/52.85 & 0.004 & x & 0.457 & 100\% & 100\% \\ 
& {\namemodel} (Transformer) & 0.7827 & 0.5871 & 0.06490 & 0.2677 & 25.58/26.90 & 35.99/ 52.85 & 0.0177 & x & 0.6508 & 96.28\% & 97.38\% \\
\hline
\end{tabular}
}}

}\caption{\label{tab:trans_lstm} Comparison of Transformer based model with LSTM(default) based modeling.}
\end{table}

\section{ Similarity of Target Dataset with Auxiliary Datasets and Impact on Performance }
\label{app:sim_tar_aux}

\begin{table}[h!]
\centering
\small
\scalebox{0.9}{
{\scriptsize
\begin{tabular}{c|c|c|c|c|c|c|c|c|c|c}
\textbf{Target dataset} & \textbf{Auxiliary Dataset} & \textbf{Deg.} & \textbf{Clus.} & \textbf{Orbit} & \textbf{NSPDK}  & \makecell{\textbf{Avg} \# \textbf{Nodes} \\ \textbf{(Target/Aux)}} & \makecell{\textbf{Avg} \# \textbf{Edges} \\ \textbf{(Target/Aux)}} & \makecell{\textbf{Node} \\ \textbf{Label}} & \makecell{\textbf{Edge} \\ \textbf{Label}} & \makecell{\textbf{Joint}  \textbf{Node} \\ \textbf{Label} \\ \& \textbf{Degree}}  \\
\hline
\multirow{8}{4em}{Leukemia-Active} & \textbf{Breast} & 0.007 & 0.0008 & 0.0003 & 0.07 & 47.0/55.83 & 49.61/59.35 & 0.007 & $\approx$ 0 & 0.029 \\
 & \textbf{Lung} & 0.001 & $\approx$ 0 & 0.0002 & 0.065 & 47.0/40.0 & 49.61/41.82 & 0.012 & $\approx$ 0 & 0.05 \\
 & \textbf{Yeast} & 0.0044 & $\approx$ 0 & $\approx$ 0 & 0.07 & 47.0/32.70 & 49.61/34.03 & 0.02 & 0.002 & 0.059 \\
 & Enzyme EC1 & 1.41 & 1.28 & 0.16 & 0.47 & 47.0/29.73 & 49.61/58.52 & 1.85 & x & 1.15 \\
 & Enzyme EC2 & 1.33 & 1.22 & 0.15 & 0.49 & 47.0/32.88 & 49.61/63.16 & 1.91 & x & 1.36 \\
 & Enzyme EC3 & 1.36 & 1.28 & 0.25 & 0.47 & 47.0/28.91 & 49.61/56.50 & 1.87 & x & 1.16 \\
 & Enzyme EC4 & 1.39 & 1.18 & 0.22 & 0.48 & 47.0/37.47 & 49.61/74.00 & 1.90 & x & 1.25 \\
 & Enzyme EC5 & 1.23 & 1.12 & 0.10 & 0.46 & 47.0/27.61 & 49.61/51.17 & 1.89 & x & 1.09 \\
 & Enzyme EC6 & 1.28 & 1.31 & 0.09 & 0.48 & 47.0/30.49 & 49.61/60.01 & 1.90 & x & 1.34 \\
\hline
\multirow{8}{4em}{Enzyme EC3}  & \textbf{Enzyme EC1} & 0.042 & 0.20 & 0.03 & 0.055 & 28.35/39.85 & 53.89/70.63 & 0.007 & x & 0.08 \\
 & \textbf{Enzyme EC2} & 0.024 & 0.09 & 0.04 & 0.055 & 28.35/28.44 & 53.89/70.63 & 0.029 & x & 0.12 \\
 & \textbf{Enzyme EC4} & 0.021 & 0.11 & 0.008 & 0.050 & 28.35/37.79 & 53.89/74.27 & 0.012 & x & 0.047 \\
 & \textbf{Enzyme EC5} & 0.007 & 0.088 & 0.008 & 0.056 & 28.35/33.6 & 53.89/64.33 & 0.024 & x & 0.05 \\
 & \textbf{Enzyme EC6} & 0.031 & 0.42 & 0.05 & 0.054 & 28.35/30.33 & 53.89/59.99 & 0.019 & x & 0.096 \\
 & Yeast & 1.39 & 1.28 & 0.27 & 0.47 & 28.35/34.63 & 53.89/56.09 & 1.86 & x & 1.11 \\
 & Breast & 1.37 & 1.287 & 0.26 & 0.49 & 28.35/44.01 & 53.89/46.26 & 1.89 & x & 1.15 \\
 & Lung & 1.39 & 1.288 & 0.271 & 0.46 & 28.35/36.63 & 53.89/38.40 & 1.875 & x & 1.121 \\
\hline
\end{tabular}
}}\caption{Distance of target datasets with different auxiliary datasets on a variety of distance metrics\label{dist_target_aux}. The  auxiliary datasets highlighted in bold are highly similar to the target datasets. }
\end{table}
We compare our target datasets specifically Enzyme EC3 and Leukemia-Active with a variety of auxiliary datasets on different graph distance metrics in Table~\ref{dist_target_aux} below. For instance, we observe that Leukemia-Active target dataset has more similarities with chemical datasets such as Breast, Lung and Yeast. Additionally, we observe it is highly dis-similar with datasets belonging to Enzyme(Biological) category. Similarly, in Table ~\ref{dist_target_aux} below we observe that Enzyme EC3 dataset is highly similar to other Enzyme datasets and dis-similar to datasets such as Breast, Lung and Yeast. 

Further, to understand the impact of choosing different auxiliary datasets based upon their similarity to target dataset, we train \namemodel with different sets of auxiliary datasets. In Table ~\ref{perf_target_aux}, we observe \namemodel convincingly performs better when auxiliary datasets that are similar to target datasets are chosen for meta-training. Specifically, \namemodel trained with Yeast,Breast and Lung as auxiliary datasets significantly outperforms \namemodel meta-trained with Enzyme datasets, when the target dataset is Leukemia-Active. Similar conclusion is observed on Enzyme EC3 dataset in Table ~\ref{perf_target_aux}.

\begin{table}[t!]
\centering
\small
\scalebox{0.8}{
{\scriptsize
\begin{tabular}{c|c|c|c|c|c|c|c|c|c|c|c|c|c}

\textbf{Target dataset} & \textbf{Auxiliary Dataset}& \textbf{Model} & \textbf{Deg.} & \textbf{Clus.} & \textbf{Orbit} & \textbf{NSPDK}  & \makecell{\textbf{Avg} \# \textbf{Nodes} \\ \textbf{(Gen/Gold)}} & \makecell{\textbf{Avg} \# \textbf{Edges} \\ \textbf{(Gen/Gold)}} & \makecell{\textbf{Node} \\ \textbf{Label}} & \makecell{\textbf{Edge} \\ \textbf{Label}} & \makecell{\textbf{Joint} \\ \textbf{Node} \\ \textbf{Label} \\ \& \textbf{Degree}} & \textbf{Novelty} & \textbf{Uniqueness}  \\ \hline
Leukemia-Active & Yeast, Breast, Lung & \namemodel & ${0.0069}$ & $\approx$ 0 & $\approx$ 0 & 0.032 & 42.35/47.71 & 44.33/50.37 & 0.0011 & $\approx$ 0 & 0.24 & 100\% & 100\% \\
 & Enzyme(1,2,3,4,5,6) & \namemodel & 0.07 & 0.005 & 0.001 & 0.20 & 39.38/47.71 & 41.32/50.37 & 0.022 & $\approx$ 0 & 1.08 & 100\% & 100\% \\
\hline
Enzyme EC3 & Enzyme(1,2,4,5,6) & \namemodel & 0.45 & 0.47 & 0.025 & 0.16 & 24.5/26.90 & 37.69/52.85 & 0.004 & x & 0.457 & 100\% & 100\% \\
 & Yeast, Lung, Breast & \namemodel & 0.54 & 0.62 & 0.029 & 0.21 & 22.89/26.90 & 34.33/52.85 & 0.007 & $\approx 0$ & 0.62 & 99.8\% & 99.8\% \\
\hline
\end{tabular}
}}
\caption{ \label{perf_target_aux}Impact of different auxiliary datasets on quality of \namemodel. We observe that when auxiliary datasets are similar to target dataset(as per table~\ref{dist_target_aux}), the performance of \namemodel is significantly better.}
\end{table}



\end{document}